\documentclass[journal]{IEEEtran}
\usepackage{latexsym}
\usepackage{amsfonts,amssymb,amsmath}
\usepackage{hyperref}
\usepackage{graphicx}
\usepackage{multirow}
\usepackage{longtable}
\usepackage{supertabular}
\usepackage{float}
\usepackage{stfloats}
\usepackage{cite}
\usepackage{subfigure}
\usepackage{diagbox}
\usepackage[edges]{forest}
\usepackage{tikz}
\usetikzlibrary{arrows,shapes,positioning,shadows,trees}
\tikzstyle{every node}=[draw=black,anchor=west,rounded corners=0.05cm,font=\footnotesize,drop shadow]
\usepackage[ruled,linesnumbered]{algorithm2e}
\usepackage{hyperref}
\hypersetup{colorlinks, citecolor=green, filecolor=pink, linkcolor=red, urlcolor=blue }

\def\BibTeX{{\rm B\kern-.05em{\sc i\kern-.025em b}\kern-.08em T\kern-.1667em\lower.7ex\hbox{E}\kern-.125emX}}
\begin{document}

\title{Compacting Deep Neural Networks for Internet of Things: Methods and Applications}

\author{Ke Zhang,~\IEEEmembership{Member, IEEE}, Hanbo Ying, Hong-Ning Dai,~\IEEEmembership{Senior Member, IEEE}, Lin Li, Yuangyuang Peng, Keyi Guo, Hongfang Yu,~\IEEEmembership{Member, IEEE}
\thanks{This work was supported in part by the Sichuan Science and Technology Program under Grant 2019YFG0405, in part by the Project of Science and Technology on Electronic Information Control Laboratory, in part by the Joint Key Research and Development Project between Sichuan and Chongqing under Grant cstc2020jscx-cylhX0004, in part by the Macao Science and Technology Development Fund under Macao Funding Scheme for Key R \& D Projects under Grant 0025/2019/AKP.}
\thanks{K. Zhang and Y. Peng are with the School of Computer Science and Engineering, University of Electronic Science and Technology of China, Chengdu, China, and Science and Technology on Electronic Information Control Laboratory, Chengdu, China. (e-mail: kezhang@uestc.edu.cn; pengyuanyuan@std.uestc.edu.cn)}
\thanks{H. Ying, L. Li and H. Yu are with the School of Communication and Information Engineering, University of Electronic Science and Technology of China, Chengdu, China. (e-mail:\{201852011823, 201822010425\}@std.uestc.edu.cn, yuhf@uestc.edu.cn)}
\thanks{H.-N. Dai is with Faculty of Information Technology, Macau University of Science and Technology, Macau, China. (e-mail: hndai@ieee.org)}
\thanks{K. Guo is with the Courant Institute of Mathematical Science, New York University, New York, USA. (e-mail:keyi.guo@nyu.edu)}
\thanks{Copyright (c) 2021 IEEE. Personal use of this material is permitted. However, permission to use this material for any other purposes must be obtained from the IEEE by sending a request to pubs-permissions@ieee.org.}
}

\IEEEtitleabstractindextext{%
\begin{abstract}
Deep Neural Networks (DNNs) have shown great success in completing complex tasks. However, DNNs inevitably bring high computational cost and storage consumption due to the complexity of hierarchical structures, thereby hindering their wide deployment in Internet-of-Things (IoT) devices, which have limited computational capability and storage capacity. Therefore, it is a necessity to investigate the technologies to compact DNNs. Despite tremendous advances in compacting DNNs, few surveys summarize compacting-DNNs technologies, especially for IoT applications. Hence, this paper presents a comprehensive study on compacting-DNNs technologies. We categorize compacting-DNNs technologies into three major types: 1) network model compression, 2) Knowledge Distillation (KD), 3) modification of network structures. We also elaborate on the diversity of these approaches and make side-by-side comparisons. Moreover, we discuss the applications of compacted DNNs in various IoT applications and outline future directions.
\end{abstract}

\begin{IEEEkeywords}
Deep Learning, Deep Neural Networks, Internet-of-Things, Model Compression.
\end{IEEEkeywords}}

\maketitle
\IEEEdisplaynontitleabstractindextext
\IEEEpeerreviewmaketitle

\section{Introduction}

\IEEEPARstart{W}{e} have experienced the proliferation in the Internet of Things (IoT), which has been widely deployed in diverse industrial and economic sectors to connect \emph{things}, \emph{people}, and \emph{processes} to form a cyber-physical-social system~\cite{noauthor_what_nodate}. Massive data has been generated from various IoT devices and IoT networks. The IoT data has characteristics of diversity, heterogeneity, and massive volumes~\cite{hndai:BDAWireless2019}. The data analytics on enormous IoT data can extract high economic and social values. However, it is challenging to collect, process, store, and analyze massive IoT data with diverse types (e.g., unstructured, structured, text, and video data).

The recent advances in computing, big data, and Artificial Intelligence (AI) bring opportunities to address data analytics challenges in IoT. In particular, cloud computing serves as a new computing paradigm to outsource IoT data to remote cloud servers, which can store and process massive IoT data. Meanwhile, distributed computing models such as MapReduce, Hadoop, and Spark can distribute computing and storage tasks to different computing nodes to balance the load. Moreover, AI-enabled data analytical methods, such as Machine Learning (ML) and Deep Learning (DL) algorithms, can analyze massive IoT data and extract valuable information.

ML approaches that are beneficial to handle structured and well-labeled data has been widely used in business analysis and decision sciences. At the same time, they cannot well process unstructured or unlabelled data. In contrast, DL algorithms based on Artificial Neural Networks (ANNs) have superior performance than conventional ML methods by mimicking human brains, which have the strengths to analyze complex data. Moreover, DL approaches can well process unstructured data without extensive human interventions. 

DL has dramatically changed the way of computing facilities processing various information. In IoT, the massive amount of heterogeneous IoT data undoubtedly provides a stage for DL to demonstrate its strength. Applying DL algorithms to IoT devices can provide users with vital services, such as intelligent transportation systems, smart manufacturing, traceable logistics, and social networks.


However, DL algorithms often spend a long time in training DL models via learning from massive data. Moreover, DL algorithms also require being executed at dedicated computing facilities, such as the Graphics Processing Unit (GPU) and Tensor Processing Unit (TPU), to shorten the training time and achieve high performance. Furthermore, DL models often have bulky size, thereby leading to extra storage and computational costs. The stringent computing and storage requirements prevent DL approaches from the wide adoption in IoT scenarios, in which IoT devices often have limited computing capability and storage capacity. 

Therefore, it is a necessity to investigate the portable DL approaches, which have less stringent computing and storage requirements on IoT devices. There are numerous studies working toward designing lightweight DL approaches. Because DL approaches are mainly based on Deep Neural Networks (DNNs), many of the recent advances are based on compacting neural networks, Knowledge Distillation (KD), and modification of network structures. Despite the advent in compacting DL models, there are few surveys on summarizing the studies on designing lightweight DL approaches, especially for IoT scenarios. For example, Han et al.~\cite{han2015deep} discussed several different approaches for compacting DNNs. However, they did not consider the advent of KD and modification of network structures, both of which can significantly compact DNN models.

\subsection{Motivation}
There are several recent surveys on DNN compression. The work~\cite{YCheng:SPM18} presents an overview of model compression and acceleration of DNNs while it only considers Convolutional Neural Networks (CNNs) and does not consider other types DNNs, such as Recurrent Neural Networks (RNNs). Meanwhile, \cite{mishra_survey_2020} provides a more detailed overview of model compression and acceleration techniques with a more detailed elaboration of each type of method. Moreover,~\cite{choudhary_comprehensive_2020} mainly addresses the compression of ML models while having a simple discussion on RNN compression. Furthermore,~\cite{deng_model_2020} investigates the impact of hardware on model compression, which nevertheless has not appeared in the previous surveys. Although these surveys address some issues on model compression and acceleration of DNNs, they seldom address any issues on practical applications. There is only one recent work~\cite{ferrag_deep_2020} discussing related applications of DL in the network security perspective while no addressing the comprehensive IoT ecosystem. 

It becomes an inevitable trend to apply DL approaches to IoT applications due to the strengths of DL. Practical applications in IoT also pose challenges on model compression and acceleration, especially distributed IoT with consideration resource allocation among multiple models. Moreover, the diversified applications of IoT also bring challenges in deploying compressed DNN models. However, most of existing studies lack of comprehensive analysis on compacting-DNNs techniques with consideration of IoT applications. 

\begin{table}[t]
    \centering
    \caption{Comparison of this paper with representative surveys}
    \renewcommand{\arraystretch}{1.25}
    \label{tab:table1-1}
\footnotesize
    \begin{tabular}{c|cccccc}
    \hline
    \diagbox{\bf Issues}{\bf Refs.}&\begin{tabular}[c]{@{}c@{}}\cite{YCheng:SPM18}\\ 2018\end{tabular} &\begin{tabular}[c]{@{}c@{}}\cite{mishra_survey_2020}\\ 2020\end{tabular} &\begin{tabular}[c]{@{}c@{}}\cite{choudhary_comprehensive_2020} \\2020\end{tabular} &\begin{tabular}[c]{@{}c@{}}\cite{deng_model_2020}\\2020\end{tabular} &\begin{tabular}[c]{@{}c@{}}\cite{ferrag_deep_2020}\\ 2020\end{tabular} &\begin{tabular}[c]{@{}c@{}}\textbf{This} \\ \textbf{paper}\end{tabular}\\ \hline
 \hline
\begin{tabular}[c]{@{}c@{}}Compacting\\  networks\end{tabular}   & Yes                 & Yes                       & Yes                                  & Yes                    & No                      & Yes          \\ \hline
\begin{tabular}[c]{@{}c@{}}Knowledge\\  distillation\end{tabular} & Partial             & Yes                       & Yes                                  & No                     & No                      & Yes          \\ \hline
\begin{tabular}[c]{@{}c@{}}Modifying\\ structures\end{tabular}   & No                  & No                        & No                                   & Partial                & No                      & Yes          \\ \hline
\begin{tabular}[c]{@{}c@{}}Applications\\ in IoT\end{tabular}     & No                  & No                        & No                                   & No                     & Partial                 & Yes         \\ \hline
    \end{tabular}
\end{table}

\subsection{Contributions}

This research gap motivates us to present a comprehensive survey on compacting DNNs for IoT. The core contribution of this paper is to present the state-of-the-art in compacting-DNNs techniques for IoT. In contrast of existing surveys, this paper presents a comprehensive survey on three major compacting-DNNs techniques as well as holistic applications of compacted DNN models in IoT. Table~\ref{tab:table1-1} compares this paper with other existing surveys. 

\begin{figure*}[ht]
    \centering
    \includegraphics[width=0.9\textwidth]{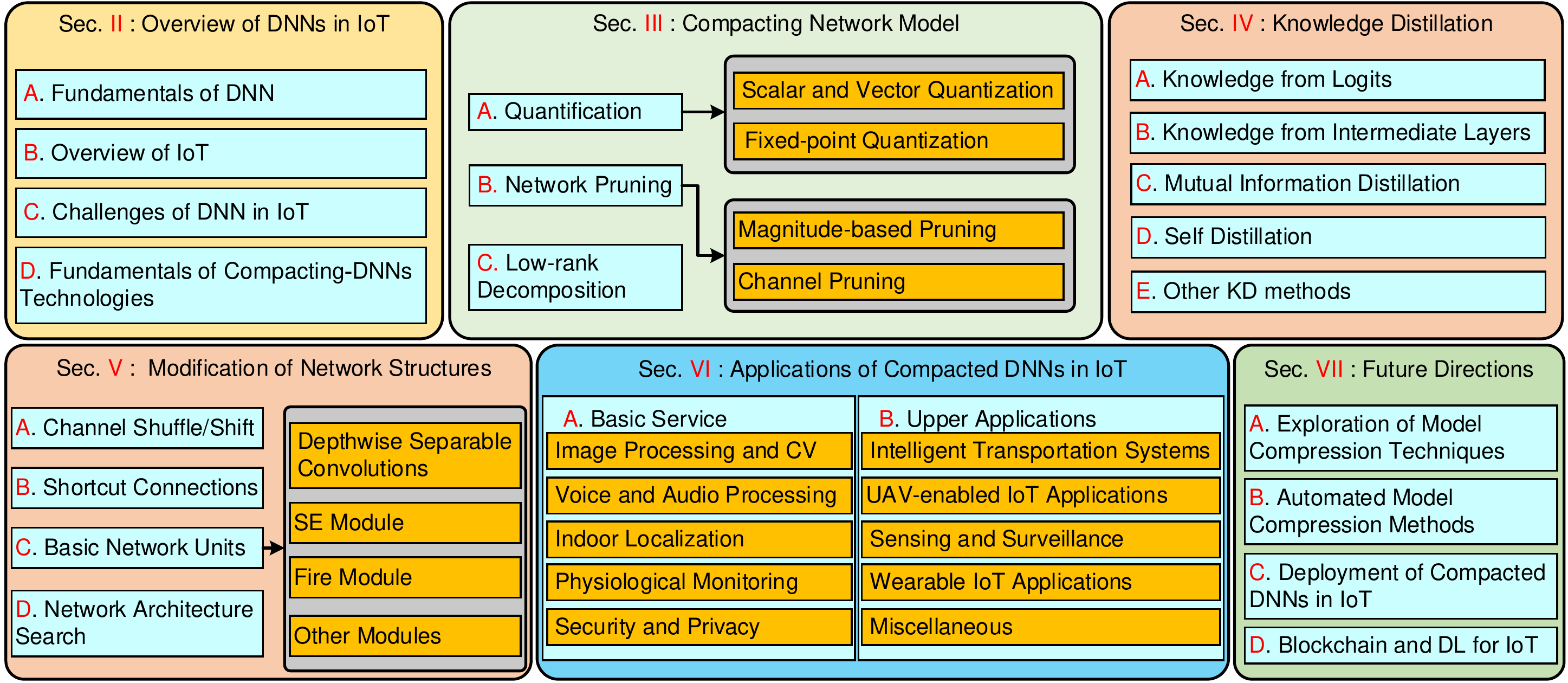}
    \caption{Diagrammatic view of the organization of this survey.}
    \label{fig1-1}
\end{figure*}

The major contributions of this paper are summarized as follows. We first present some fundamentals related to DNN and IoT in Section~\ref{sec:overview}. We then categorize the major compacting-DNNs approaches into three categories: 1) compacting network model in Section~\ref{sec:compacting}, 2) KD in Section~\ref{sec:kd}, and 3) modification of network structures in Section~\ref{sec:modif}. In each category, there are several different approaches, as shown in Fig.~\ref{fig1-1}. Unlike the previous surveys, we mainly concentrate on the breakthroughs and innovations brought by the latest advances in compacting DNNs. In terms of IoT applications, we discuss the role of DNN in different application domains as well as the role of compression and acceleration methods in Section~\ref{sec:app}. We also outline the future directions of this field in Section~\ref{sec:future}. We finally conclude the paper in Section~\ref{sec:conc}. The abbreviations of important terms are given in Appendix~\ref{sec:abbr}.

\section{Overview of DNNs in IoT}
\label{sec:overview}

\subsection{Fundamentals of DNN}

AI may have come on in leaps and bounds in the last few years, but it still has a distance away from real intelligence, which is expected to reason and make decisions like humans. DL is a type of ML and AI, which imitates the way humans obtain specific knowledge. Normally, DL algorithms are stacked with increasingly complex and abstract hierarchical structures. By using DL, the process of collecting, analyzing, and interpreting large amounts of data becomes faster and easier. 

The success of DL is mainly attributed to ANNs, which consist of a number of bionic neurons similar to millions of neurons in a human brain~\cite{VSze:PIEEE17}. Neural networks mimic the way that human brains achieve perception, recognition, and inference. The early design of ANNs limits ANN to a three-layer structure: one input layer, one hidden layer, and one output layer. DNN that further extends an ANN has multiple hidden layers between the input layer and the output layer, where ``deep'' means the multiple hidden layers. Similar to ANNs, DNNs can also fit complex nonlinear relationships. DNNs can be roughly categorized into the following types: CNNs, RNNs, and other types of DNNs.

\begin{figure}[t]
    \centering
    \includegraphics[width=0.45\textwidth]{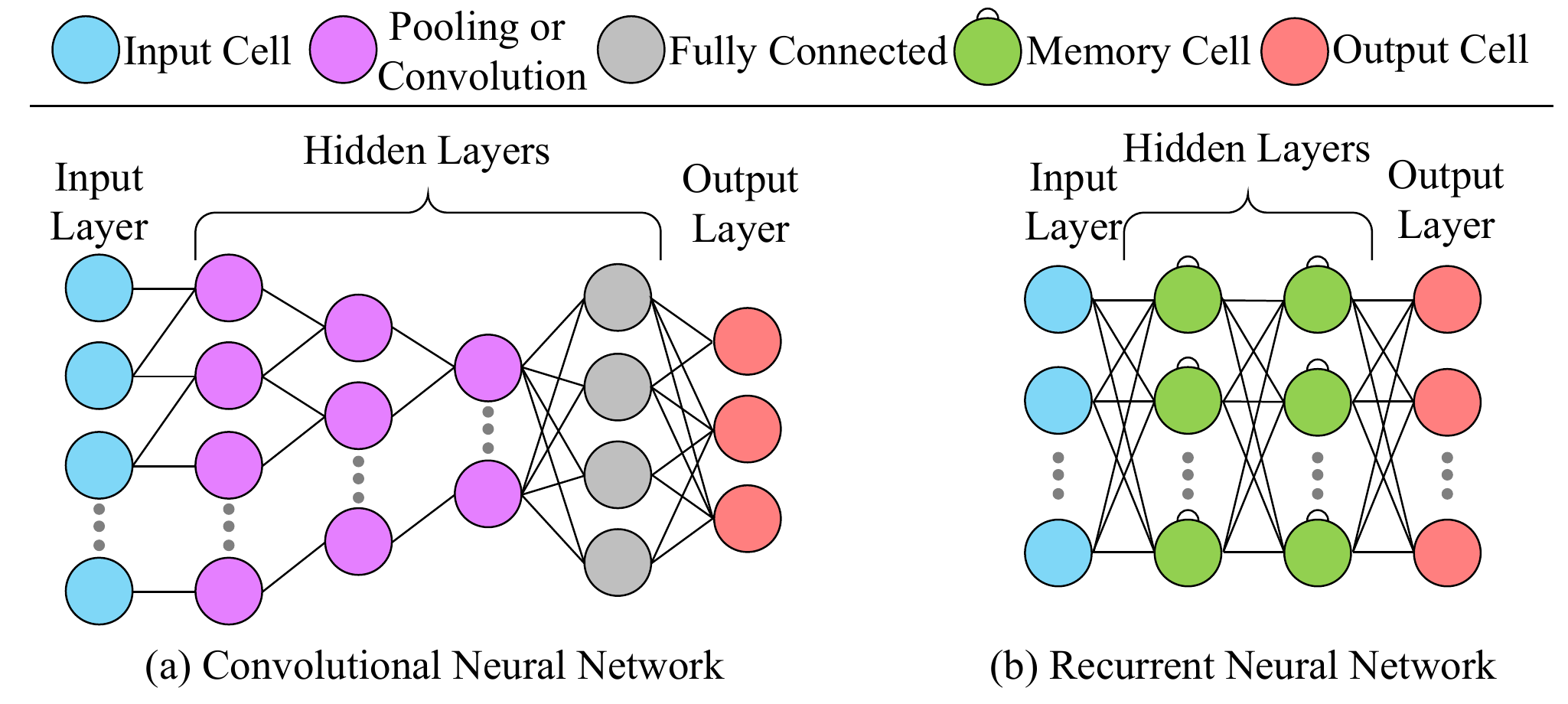}
    \caption{General structures of CNN and RNN. The main structure of CNNs consists of convolutional layers, while the main structure in DNNs consists of memory cells.}
    \label{fig2-1}
\end{figure}

\emph{1) CNN:} CNNs were initially proposed by LeCun et al.~\cite{lecun_backpropagation_1989}. The emergence of AlexNet~\cite{krizhevsky_imageNet_2017} in 2012 was regarded as the \emph{Renaissance} of neural networks and the rise of DL. CNN mainly contains convolutional layers and sub-sampling layers. In the convolutional layer, neurons only connect to some adjacent neurons, as in Fig.~\ref{fig2-1}\textcolor{red}{(a)}. This sparse connection reduces a large number of parameters and the risk of overfitting~\cite{matthew_visualizing_2014}. The convolutional layer usually contains several feature maps. Each feature map is composed of neurons with shared weights arranged in a rectangle. The convolution kernel is usually initialized with a random decimal matrix. Sub-sampling can be regarded as a special convolution process, also called pooling. Its function is to reduce the number of parameters in the feature maps and use a feature value to represent a small area. Convolution and sub-sampling work together to reduce model complexity and parameters. Generally, CNN consists of three parts: 1) input layer, 2) a combination of $n$ convolutional layers and pooling layers, 3) a fully connected multi-layer perceptron classifier. Some classic CNN structures include AlexNet~\cite{krizhevsky_imageNet_2017}, Visual Geometry Group (VGG)~\cite{simonyan_very_2015}, GoogleNet~\cite{szegedy_going_2015}, and ResNet~\cite{he_deep_2016}.

\emph{2) RNN:} Many practical applications deal with different types of data such as text, voice, and video, all of which are nearly related. Moreover, the network's output is related to the input at current time and connected to one or more previous outputs at a specific time. Traditional neural networks cannot handle this temporal relation. Another problem is that both the input and output data formats of traditional neural networks are fixed. In contrast, some practical problems, such as machine translation, require variable-length input and output data. Therefore, a more robust model is needed to solve this problem. RNNs can solve this problem since a neuron in RNNs can receive the information from both the top layer and the previous RNN unit~\cite{schmidhuber_deep_2015}, as in Fig.~\ref{fig2-1}\textcolor{red}{(b)}. Commonly-used RNN methods are multi-layer RNN, bidirectional RNN, and recursive neural network. Other variants of RNNs include Long Short-Term Memory (LSTM)~\cite{hochreiter_long_1997} and Gated Recurrent Unit (GRU).~\cite{cho_learning_2014}.

\emph{3) Other DNNs:} DNN has some other novel network structures while we only introduce the two most representatives of them. Hinton proposed Deep Belief Networks (DBNs)~\cite{hinton_reducing_2006} in 2006. A DBN is a generative model that can express an in-depth representation of training data and its structure, consisting of multiple restricted Boltzmann machines. DBNs are usually used as a pre-trained component of other DNNs and consequently form many neural networks based on DBNs. The typical example is convolutional DBNs. The Generative Adversarial Network (GAN)~\cite{goodfellow_generative_2014} proposed by Goodfellow is also a variant of DNN. A GAN is composed of two models: a discriminative model and a generative model. The responsibility of the discriminative model is to differentiate the real data and the created data accurately. The generative model's role is to generate new data that is sufficiently similar to the real data. At present, GANs mainly devote to computer vision and natural language processing, such as improving image resolution, restoring occlusion images, and generating images based on text descriptions. The literature~\cite{NOWROOZI2021102092} introduced GANs in adversarial-image forensics.

\subsection{Overview of IoT}

The term ``Internet of Things'' was first proposed by Ashton in 1999 to describe which a system can connect objects attached to sensors in the physical world to the Internet~\cite{noauthor_internet_nodate}. Today, IoT describes a network of physical objects embedded in some components, with the purpose of data connection and exchange with other devices and systems over the Internet. Nowadays, there are more than 7 billion connected IoT devices. And experts predict that this number will grow to 10 billion by 2020 and 22 billion by 2025~\cite{noauthor_what_nodate}.

The proliferation of IoT devices leads to massive IoT data generated all the time. This volume of IoT data contains valuable information as well as noise, error, and redundant information~\cite{hndai:BDAWireless2019}. It is necessary to apply sophisticated ML methods to process and analyze IoT data. How to reliably obtain actual IoT data from a noisy and complex environment is a problem. It is not enough to gather a massive amount of IoT data, and it is also essential to implement data analysis~\cite{verma_survey_2017}. The analytics on the IoT data can be categorized into 1) stream analytics and 2) real-time analytics~\cite{akbar_predicting_2015}. The former has no time limit or average time limit while the latter needs to provide analysis response or output within a strict time limit when the IoT data reaches the micro-batches~\cite{mohamed_real-time_2014}.

DL is an effective method that can extract useful information from massive IoT data. Its high efficiency in complex data processing may play an essential role in future IoT services. Compared with ML, DL has a better performance with a massive amount of data. Furthermore, it also can automatically extract new features for different problems~\cite{li_learning_2018}. In IoT, many researchers have used DNN to simplify complex problems. For example, the handover control in wireless systems via DNN~\cite{wang_handover_2018}, DNN-based complex resource allocation solution for the collaborative mobile edge computing network~\cite{chen_iraf_2019}, beam management and interference coordination with DNN in dense millimeter wave networks~\cite{zhou_deep_2019}, and the use of a DNN to speed up resource allocation in wireless networks~\cite{lee_learning_2020}. In the area of the smart building~\cite{zhang_thermal_2019}, DNN is utilized for thermal comfort modeling. Also, in the field of Internet of Vehicles (IoV)~\cite{song_blockchain_2020} there is a blockchain-enabled IoV framework with cooperative positioning methods to improve the positioning accuracy, system robustness, and security of vehicular Global Positioning System (GPS). With the introduction of Fifth Generation (5G) networks, more and more devices will join the network in the future. In a massive device communication scenario, most devices spend most of their time in the sleep state and seldomly need to remain active to save energy. Therefore, to ensure the efficiency and success of communication, it is necessary to monitor the equipment's status. A DNN based on variational autoencoder was built to detect device activity in massive machine-type communications under imperfect channel state information~\cite{zhao_a_2020}.

\subsection{Challenges of DNNs in IoT}

There is no doubt that the advantages of DNNs and IoT complement each other. However, how to effectively combine DNNs and IoT is a problem that many scientific researchers attempt to solve. Cloud computing~\cite{zhang_cloud_2010} has some inherent advantages such as virtualization, large-scale integration, high reliability, high scalability, and relatively low cost. It comprises three key service models: 1) Infrastructure-as-a-Service (IaaS), 2) Platform-as-a-Service (PaaS), and 3) Software-as-a-Service (SaaS)~\cite{mouradian_a_2018}. Because of enormous DNN models and computational complexity, it is difficult to compute the inference results at devices with limited resources. The intuitive solution is to deploy the DNN model on the server in the cloud data center. However, with so many computing tasks being processed in the cloud, the data that needs to be transmitted is large in scale and quantity, putting tremendous pressure on the cloud computing infrastructure's network capacity and computing power. Besides, almost all applications in IoT field require ultra-low power consumption, little storage space consumption, and real-time data processing, especially for those sensitive to delay or highly interactive. The considerable amount of data communication has dramatically increased the pressure on the backbone network and brought massive expansion and maintenance costs to service providers. In order to cope with the excessive resource requirements of DNNs, traditional methods rely on mighty clouds for DNN calculations. However, using this method may have severe delays and wasted energy~\cite{li_edge_2020}. IoT applications have vital real-time requirements and user data privacy issues in actual use. It is more efficient and secure to deploy DNNs at edge computing nodes instead of remote cloud servers. 

Fog computing~\cite{flavio_fog_2012} and edge computing~\cite{shi_edge_2016} are attractive supplements to cloud computing. The basic idea of fog/edge computing is to deploy computing facilities close to the data source for data processing, rather than transmitting data to a remote computing facility. In edge computing, the vast majority of occasions are mobile edge computing~\cite{dahmen-lhuissier_etsi_nodate}. Fog/edge computing-based IoT is a distributed architecture, which is physically close to the location where data is generated, thereby making it outstanding in service delivery and privacy~\cite{omoniwa_fog_2019}. This architecture consists of two layers: the edge layer and the cloud layer. The former layer usually consists of sensors, vehicles, various IoT devices, IoT gateways, and network access points. The cloud layer contains internet connections and cloud servers. In this way, not all the data is transmitted to remote clouds, thereby reducing data transmission. 

Because distributed processing through edge computing or parallel computing through mobile devices will increase additional transmission costs and energy consumption, Umeda and Karin~\cite{umeda_processing_2019} proposed a way to divide DL tasks of immediate layers (hidden layers) based on node processing power. They distributed a hidden layer to several nodes in a wireless sensing network and assigned their tasks according to their computing power in advance. Then each node performs tasks in sequence, and the output from the previous node will be considered as the input of the next node. If some nodes are missing, the remaining tasks will be handed over to the server. The combination of edge computing and AI has been a new hot area, being referred to as ``Edge Intelligence (EI)'' or ``edge AI''~\cite{wang_in-edge_2019}. Specifically, according to the amount and path length of data offloading, EI is categorized into six levels. The definition is given as follows~\cite{zhou_edge_2019}: 1) Cloud Intelligence, 2) Cloud–Edge Conference and Cloud Training, 3) In-Edge Co-inference and Cloud Training, 4) On-Device Inference and Cloud Training, 5) Cloud–Edge Co-training and Inference, 6) All In-Edge, 7) All On-Device.

\begin{figure*}[ht]
    \centering
    \includegraphics[width=0.9\textwidth]{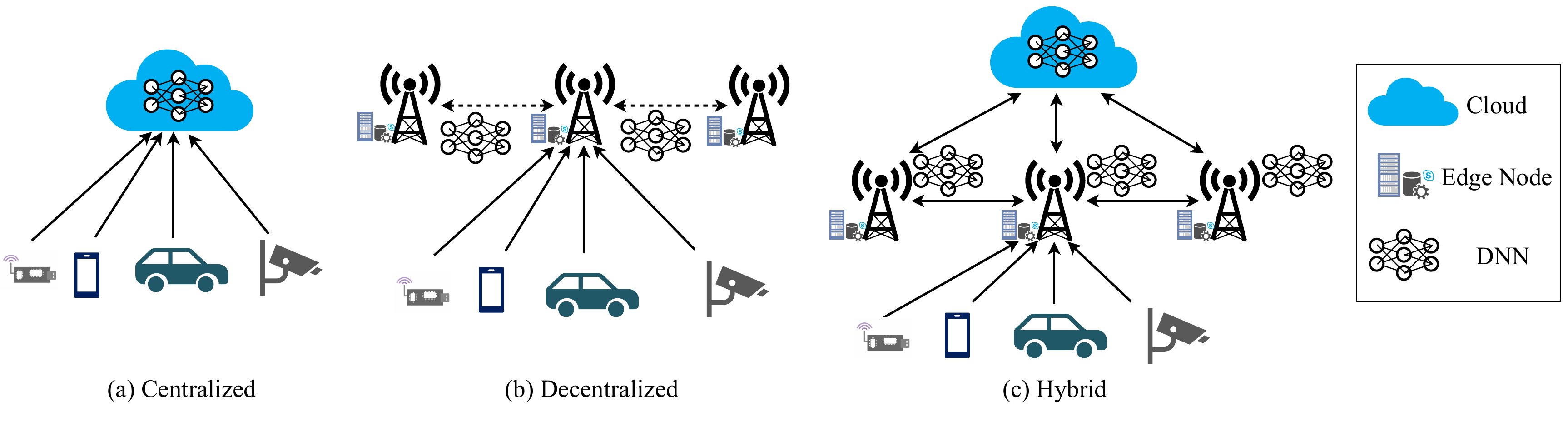}
    \caption{DNN model deployment in IoT. (a) The centralized computing tasks are undertaken by the clouds; (b) the decentralized computing tasks are undertaken by the edge nodes; (c) the hybrid computing tasks are shared by the clouds and edge nodes after being appropriately allocated.}
    \label{fig2-2}
\end{figure*}

There are three ways, as shown in Fig.~\ref{fig2-2}, to deploy DNNs in IoT: 1) centralized mode, 2) decentralized mode, and 3) hybrid mode.

\emph{(a) Centralized:} In this mode, the training and inference tasks of DNN models are performed at cloud servers. The training data is generated and collected directly from the end devices. The training data can be a collated dataset, which can be directly uploaded to remote clouds. However, this mode has a high delay and is only suitable for applications that do not have vital real-time requirements.

\emph{(b) Decentralized:} In this mode, edge computing nodes (often deployed with IoT gateways and base stations) can collect IoT data to train the DNN model locally, thereby protecting private information. The global DNN model can be obtained by sharing the local DNN models between the edge servers. The end devices can download the models from the local edge computing nodes or download them from other edge nodes.

\emph{(c) Hybrid:} The configuration in this mode is the most flexible. In this case, edge computing nodes are distributed throughout the entire architecture. The DNN models can be integrated through distributed updating or centralized training in the cloud center.
 
The mixed-mode is the most widely used, but this mode's resource scheduling problem is also the most complicated. How can the scheduling task achieve the response with the lowest possible delay? Lin et al.~\cite{lin_cost-driven_2020} proposed a self-adaptive particle swarm optimization algorithm with genetic algorithm operators, which cuts down the cost of data transmission as well as execution and controls the expiry date of applications. However, this method does not consider the load-balancing problem.

IoT devices are usually limited by resources, which leads to the need for distributed DNN models to be deployed across IoT devices. Regarding the distributed design of DNN deployment, the existing solutions are categorized into two categories: 1) focuses on how to compress the DNN model to reduce the amount of calculation, and 2) focuses on how to partition the DNN model. Thus, each device is only responsible for the partial computation, it can afford~\cite{chang_an_2019}. DNN models usually have a very deep structure with many parameters. So, they often have significant redundancy in the parameterization of several DL models~\cite{denil_predicting_2013}. Thus, recent studies such as~\cite{lei_do_2014} attempt to convert a deep model to a shallow neural network even with a similar number of parameters. In this paper, we mainly investigate the methods of compressing DNNs through various techniques.

\subsection{Fundamentals of Compacting-DNNs Technologies}
As mentioned above, it is necessary to compact existing DNNs in terms of the model size and the computing cost so that compacted DNNs can be used in different IoT applications. In particular, compacting-DNNs technologies aim to solve model efficiency, which mainly focuses on the storage cost (i.e., the model size) and the time cost (i.e., the computing cost) during inference phase of the model. The storage problem mainly refers to the storage of weight parameters in the model, which requires enormous footprints for the device. The time problem mainly refers to the number of computations.

Generally, the methods of compacting DNNs can be categorized into two categories: 1) compressing the original model into a compact model and 2) directly training a small-size model. The main idea of the first group of methods is to decrease the parameters and computation tasks in a pre-trained model, typically including quantization, network pruning, and low-rank decomposition. The second group of methods includes a) KD and b) modification of network structures. KD mainly allows a student network to learn and imitate the output of an excellent teacher network as the teacher network's output contains more information than the original data labels so that the student network with a much smaller size can have a similar performance to the teacher. Modification of network structures mainly relies on Network Architecture Search (NAS) and artificial basic unit design, which can find small but efficient and representative basic units. According to specific application scenarios, the modified model can be scaled up and fine-tuned appropriately, and then it can be put into use.

\section{Compacting Network Model}
\label{sec:compacting}

From the perspective of lightweight network structures, the original bloated layer structures can be reduced and streamlined through quantization, pruning and low-rank decomposition techniques.

\subsection{Quantization}

Quantization is a method for many models to compress and accelerate applications. The goal of quantization is to reduce the overhead of high-precision floating-point calculations. The earliest quantization used are Product Quantization (PQ)~\cite{jegou_product_2011} and Residual Quantization (RQ)~\cite{chen_approximate_2010}. According to the quantification of objects, quantitative methods have two categories: 1) scalar and vector quantization and 2) fixed-point quantization.

\subsubsection{Scalar and Vector Quantization.}

By using these techniques, a codebook and a set of quantization codes represent the original parameters. The quantization codes describe the distribution of quantization centers in the codebook. If further compression is needed, the quantization code can also be encoded by lossless encoding (e.g., Hoffman encoding). In Table~\ref{tab:table3-1}, we briefly compare some scalar and vector quantization methods.

\begin{table}[t]
  \scriptsize
  \centering
\renewcommand{\arraystretch}{1.25}
  \caption{Comparison of scalar and vector quantization methods according to quantization algorithm and encoding method.}
  \label{tab:table3-1}
  \begin{tabular}{c|cc}
  \hline
  \textbf{References} &
    \textbf{Quantization algorithm} &
    \textbf{Encoding method} \\ \hline
  Gong et al.~\cite{gong_compressing_2014} &
    \begin{tabular}[c]{@{}c@{}}$k$-means quantization,\\ PQ, RQ\end{tabular} &
    N/A \\ \hline
  Wu et al.~\cite{wu_quantized_2016} &
    \begin{tabular}[c]{@{}c@{}}PQ and sharing \\ block weights\end{tabular} &
   N/A \\ \hline
  Gudovskiy et al.~\cite{gudovskiy_shiftcnn_2017} &
    similar to RQ &
    N/A \\ \hline
  Choi et al.~\cite{choi_towards_2017} & \begin{tabular}[c]{@{}c@{}}Hessian weighted \\ $k$-means quantization\end{tabular} & Huffman codes \\ \hline
  Reagen et al.~\cite{reagen_weightless_2017} &
    N/A &
    \begin{tabular}[c]{@{}c@{}}Based on the \\ Bloomier filter\end{tabular} \\ \hline
  \end{tabular}
\end{table}

In~\cite{gong_compressing_2014}, Gong et al. proposed the vector quantization method and explored scalar and vector quantization technology. The common methods of scalar and vector quantization are $k$-means quantization, PQ, and RQ. The $k$-means quantization works as follows: given the parameter $W \in R^{m \times n}$, all scalar values can be collected as $w \in R^{1 \times m n}$, and the following values can be obtained by the $k$-means algorithm:

\begin{equation}\min \sum_{i}^{m n} \sum_{j}^{k}\left\|w_{i}-c_{j}\right\|_{2}^{2},\end{equation}where $w$ and $c$ are scalars. Once clustering is done, all $w$ is assigned a corresponding clustering index, and the cluster center $c^{1 \times k}$ forms a codebook. In the prediction stage, the value of each $w_{ij}$ in the codebook can be directly found. Therefore, the reconstruction matrix is:

\begin{equation}\widehat{W}_{i j}=c_{z}, \text { where }  \min _{z}\left\|W_{i j}-c_{z}\right\|_{2}^{2}.\end{equation}

\emph{PQ:} PQ's intuition is to partition the vector space into many disjoint subspaces and then perform quantization in each of these subspaces since it assumes that the vectors in each subspace are highly redundant. Concretely, given matrix $W$, it can be partitioned into several submatrices:
\begin{equation}W=\left[W^{1}, W^{2}, \cdots, W^{s}\right],\end{equation}where $W^i\in R^{m\times\left(n/s\right)}$, it is assumed that $n$ can be divided by $s$. Then the $k$-means clustering method is conducted to each submatrix $W^i$ as follows:
\begin{equation}\min \sum_{z}^{m} \sum_{j}^{k}\left\|w_{z}^{i}-c_{j}^{i}\right\|_{2}^{2},\end{equation}
where $w_z^i$ denotes the $z$-{th} row of sub-matrix $W^i$, $c_j^i$ denotes sub-codebook the $j$-th row of $C^i\in R^{k\times\left(n/s\right)}$. For each sub vector, only the corresponding cluster index and codebook are needed. Thus, the reconstruction matrix is as follows:
\begin{equation}\widehat{W}=\left[\widehat{W}^{1}, \widehat{W}^{2}, \cdots, \widehat{W}^{s}\right],\end{equation}
\begin{equation}\widehat{w}_{j}^{i}=c_{j}^{i}, \text { where } \min _{j}\left\|w_{z}^{i}-c_{j}^{i}\right\|_{2}^{2}.\end{equation}

\emph{RQ:} RQ's basic idea is first quantizing the vector into $k$ centers and then quantizing the residual recursively. Given a set of vectors $\boldsymbol{w}_{i}, i \in 1, \cdots, m$, they are quantized into $k$ different vectors by $k$-means clustering:

\begin{equation}\min \sum_{z}^{m} \sum_{j}^{k}\left\|\boldsymbol{w}_{z}-\boldsymbol{c}_{j}^{1}\right\|_{2}^{2}.\end{equation}For each $\boldsymbol{W}_{z}$, it is represented by its closest center $\boldsymbol{c}_j^1$. Next, the residuals between $\boldsymbol{w}_z$ and $\boldsymbol{c}_j^1$ are calculated for all data points, and the residuals vector $\boldsymbol{r}_z^1$ is recursively quantized into $k$ different code words $\boldsymbol{c}_j^2$. Finally, it can reconstruct the vector by adding its corresponding center at each stage:

\begin{equation}\widehat{\boldsymbol{w}}_{z}=\boldsymbol{c}_{j}^{1}+\boldsymbol{c}_{j}^{2}+\cdots+\boldsymbol{c}_{j}^{t}.\end{equation}It is assumed that $t$ iterations have been performed recursively.

\begin{table}[t]
    \centering
    \caption{Scalar versus vector quantization compression rates.}
    \renewcommand{\arraystretch}{1.25}
    \label{tab:table3-2}
    \begin{tabular}{c|c}
    \hline
    \textbf{Methods}         & \textbf{Compression rate} \\ \hline
    $k$-means quantization  & $32 / \log _{2}(k)$                                      \\ \hline
    PQ                      & $(32 m n) /\left(32 k n+\log _{2}(k) s m\right)$           \\ \hline
    RQ                      & $m /\left(t k+\log _{2}(k) t n\right)$                     \\ \hline
    \end{tabular}
\end{table}

The three methods' compression rates are different, and the specific compression rates are compared in Table~\ref{tab:table3-2}. Wu et al.~\cite{wu_quantized_2016} used a strategy of PQ, which allows less storage space by sharing block weights. With sharing, dot multiplication is converted to the addition operation. Gudovskiy et al.~\cite{gudovskiy_shiftcnn_2017} used a method similar to residual quantification. Choi et al.~\cite{choi_towards_2017} designed an overall network quantization scheme to minimize the loss under the proportional compression constraints. This method mainly introduces Hessian weighted $k$-means clustering method to quantify network parameters. Reagen et al.~\cite{reagen_weightless_2017} proposed a novel lossy encoding method based on the weight-free probability data structure of the Bloomier filter, which did not accurately store the function mapping. 

\subsubsection{Fixed-point Quantization}

This method is a low-precision quantification of parameters. It uses fixed-point numbers instead of floating-point numbers to fully use the advantages of fixed-point calculation over floating-point calculation, which reduces memory consumption and computational complexity. Fixed-point quantization is categorized into low-bit quantization and binary/ternary quantization. In Table~\ref{tab:table3-3}, we briefly compare some fixed-point quantization methods.

\begin{table}[htb]
    \centering
    \caption{Comparison of fixed-point quantization methods.}
    \renewcommand{\arraystretch}{1.25}
    \label{tab:table3-3}
    \begin{tabular}{c|ccc}
    \hline
    \multirow{2}{*}{\bf References} & \multicolumn{3}{c}{\bf Quantization}         \\ \cline{2-4} 
                              & \textbf{Weight}              & \textbf{Activation} & \textbf{Gradient} \\ \hline
    HWGQ~\cite{cai_deep_2017}       & Binary              & 2bit       & Full     \\ \hline
    DNQ~\cite{xu_dnq_2019}        & 3bit, 5bit, Dynamic & Full       & Full     \\ \hline
    BNN~\cite{courbariaux_binarized_2016}        & Binary              & Binary     & Full     \\ \hline
    XNOR~\cite{rastegari_xnor-net_2016}       & Binary              & Binary     & Full     \\ \hline
    QNN~\cite{hubara_quantized_2016}        & Binary              & Binary     & Full     \\ \hline
    TWN~\cite{li_ternary_2016}        & Binary, Ternary     & Full       & Full     \\ \hline
    ATN~\cite{ding_asymmetric_2017}        & Ternary             & Full       & Full     \\ \hline
    RTN~\cite{li_rtn_2019}        & Ternary             & Ternary    & Full     \\ \hline
    \end{tabular}
\end{table}

\subsubsection*{Low-bit Quantization}

There are two rounding methods for fixed-point quantization~\cite{gupta_deep_2015}: 1) round-to-nearest and 2) stochastic rounding.

The method of round-to-nearest is defined as:
\begin{equation}
\footnotesize
\mathrm{Round} \big(x,\langle\mathrm{IL}, \mathrm{FL}\rangle \big)=\left\{\begin{aligned}
    \lfloor x\rfloor, \quad & \text{if } \lfloor x\rfloor \leq x \leq \lfloor x\rfloor+\frac{\epsilon}{2}; \\
    \lfloor x\rfloor+\epsilon, \quad & \text{if } \lfloor x\rfloor+\frac{\epsilon}{2} \leq x \leq \lfloor x\rfloor+\epsilon,
\end{aligned}\right.\end{equation}where $\mathrm{IL}$ is the integer part of fixed-point numbers, $\mathrm{FL}$ is the decimal part of fixed-point numbers. Both $\mathrm{IL}$,$\mathrm{FL}$ are fixed-point numbers, $x$ is the real value variable, and $\epsilon=2^{-\mathrm{FL}}$ denote the smallest positive number.

The method of stochastic rounding is defined as:
\begin{equation}
\footnotesize
\mathrm{Round} \big(x,\langle\mathrm{IL}, \mathrm{FL}\rangle\big)=\left\{\begin{aligned}
    \lfloor x\rfloor, \quad & \text { probability } 1-\frac{x-\lfloor x\rfloor}{\epsilon}; \\
    \lfloor x\rfloor+\epsilon , \quad & \text { probability } \frac{x-\lfloor x\rfloor}{\epsilon}.
\end{aligned}\right.\end{equation}

Irrespective of the rounding mode used, if $x$ lies outside the range of $\langle\mathrm{IL}, \mathrm{FL}\rangle$, change the result to the lower or the upper limit of $\langle\mathrm{IL}, \mathrm{FL}\rangle$:
\begin{equation}\footnotesize\mathrm{Convert} \big(x,\langle\mathrm{IL}, \mathrm{FL}\rangle\big)= \\
  \left\{\footnotesize \begin{aligned}
    2^{\mathrm{IL}-1}, & \text { probability } 1 -\frac{x-\lfloor x\rfloor}{\epsilon}; \\
    2^{\mathrm{IL}-1}-2^{-\mathrm{FL}}, & \text { probability } 1-\frac{x-|x|}{\epsilon}; \\
    \operatorname{Round}(x,\langle I L, F L\rangle), & \text { otherwise}.
\end{aligned}\right.\end{equation}

Gupta et al.~\cite{gupta_deep_2015} proposed a new rounding method: stochastic rounding. When an underflow is generated, it is randomly rounded to one of the two numbers closest to it, and its probability is inversely proportional to the distance among them. Inspired by stochastic depth and dropout, Dong et al.~\cite{dong_learning_2017} proposed a stochastic quantization algorithm to overcome the problem of accuracy degradation. The half-wave gaussian quantization network that was proposed by Cai et al.~\cite{cai_deep_2017} is mainly used to quantify the activation value, which theoretically analyzes how to select an activation function and uses an approximate method to fit the quantization loss. Xu et al.~\cite{xu_dnq_2019} proposed a dynamic network quantization framework. Unlike most existing quantization methods, using the universal quantization bit width of the whole network, the authors used a strategy gradient to train agents to learn the bit width of each layer by bit width controller. Sakr et al.~\cite{sakr_accumulation_2019} proposed a statistical method to analyze the impact of the reduced cumulative accuracy on DL training. Yang et al.~\cite{yang_quantization_2019} defined low-bit quantization as a differentiable non-linear function (called quantization function) and proposed a new way to explain and implement neural network quantization.

\subsubsection*{Binary/Ternary Quantization}

Binary quantization represents the parameters with $1$ bit, which can reduce the storage and calculation cost of parameters to the greatest extent. Similar to low-bit quantization, there are two binarization methods~\cite{courbariaux_binarized_2016}: 1) definition and 2) stochastic.

The \emph{definition method} is defined as:
\begin{equation}
\footnotesize
x^{b}=\operatorname{sign}(x)=\left\{\begin{array}{cc}
    +1, & \text{if } x \geq 0 ;\\
    -1, & \text{otherwise} ,
\end{array}\right.\end{equation}
where $x^b$ is a binary variable (weight or activation) and $x$ is a real-valued variable.

The stochastic method is defined as:
\begin{equation}
\footnotesize
x^{b}=\left\{\begin{array}{ll}
    +1, & \text { probability }  p=\sigma(x); \\
    -1, & \text { probability }  1-p ,
\end{array}\right.\end{equation}
where $\sigma$ is the $\mathrm{hard sigmoid}$ function given below:
\begin{equation}\footnotesize\sigma(x)=\operatorname{cilp}\left(\frac{x+1}{2}, 0,1\right)=\max \left(0, \min \left(1, \frac{x+1}{2}\right)\right).\end{equation}

\begin{algorithm}[t]
\small
  \caption{ General network pruning algorithm.}
  \label{alg:algorithm1}
  \KwIn{Pre-trained model: $M$; Dataset for fine-tune: $D$; Criterion of network pruning: $C$.}
  \KwOut{Lightweight model of $M_{out}$.}
  
  Calculate criterion values of neurons or connections in $M$;
  
  Compare these values with $C$;
  
  Prune neurons or connections whose values are lower than $C$;
  
  Fine-tune the network $M_{out}$ on $D$.
  
\end{algorithm}

Courbariaux et al.~\cite{courbariaux_binarized_2016} quantized the weights and activations to $1$ bit simultaneously and gave the specific implementation for the internal hardware calculation. Rastegari et al.~\cite{rastegari_xnor-net_2016} binarized the weight and expressed the input as binary. Hubara et al.~\cite{hubara_quantized_2016} proposed a method to train the Quantized Neural Network (QNN). Lin et al.~\cite{lin_binarized_2017} proposed a Binarized CNN (BCNN) with a separable filter in binary quantization network, which applies Singular Value Decomposition (SVD) on BCNN kernels. Unlike simple matrix approximation, Hou et al.~\cite{hou_loss-aware_2018} proposed a proximal Newton algorithm with diagonal Hessian approximation that directly minimizes the loss w.r.t. the binarized weights.

The ternary quantization is proposed to make up for the low accuracy of binary quantization. Li et al.~\cite{li_ternary_2016} introduced the Three elements Weight Network (TWN) where the weights of the neural network were limited to $+1$, $0$, and $-1$. Ding et al.~\cite{ding_asymmetric_2017} proposed an asymmetric ternary network based on TWN. Li et al.~\cite{li_rtn_2019} proposed a new reparameterized three-element network to solve the problem of the forward squeezing behavior of the previous three-valued network and the saturation behavior of the backward quantization function.

\subsection{Network Pruning}

Network pruning is a type of model compression that helps to address overfitting and lower the complexity. The basic idea of network pruning is to cut a certain proportion of unimportant parts. The network pruning technique originated in the late 1980s. Hanson et al.~\cite{hanson_compating_1988} proposed a magnitude-based pruning method that minimizes the number of hidden units by applying a weight decay to these units. Chauvin et al.~\cite{touretzky_advances_1990} and Hassibi et al.~\cite{hassibi_optimal_1993} recommended measuring the importance of weights based on the loss function and then cropped them. Their ideas have had a profound influence on current works nowadays. The general process of the pruning algorithm is shown in Algorithm~\ref{alg:algorithm1}. We only discuss magnitude-based pruning, channel pruning, and some other types of pruning.

\subsubsection{Magnitude-based Pruning}

\begin{figure}[t] 
    \centering 
    \includegraphics[width=0.48\textwidth]{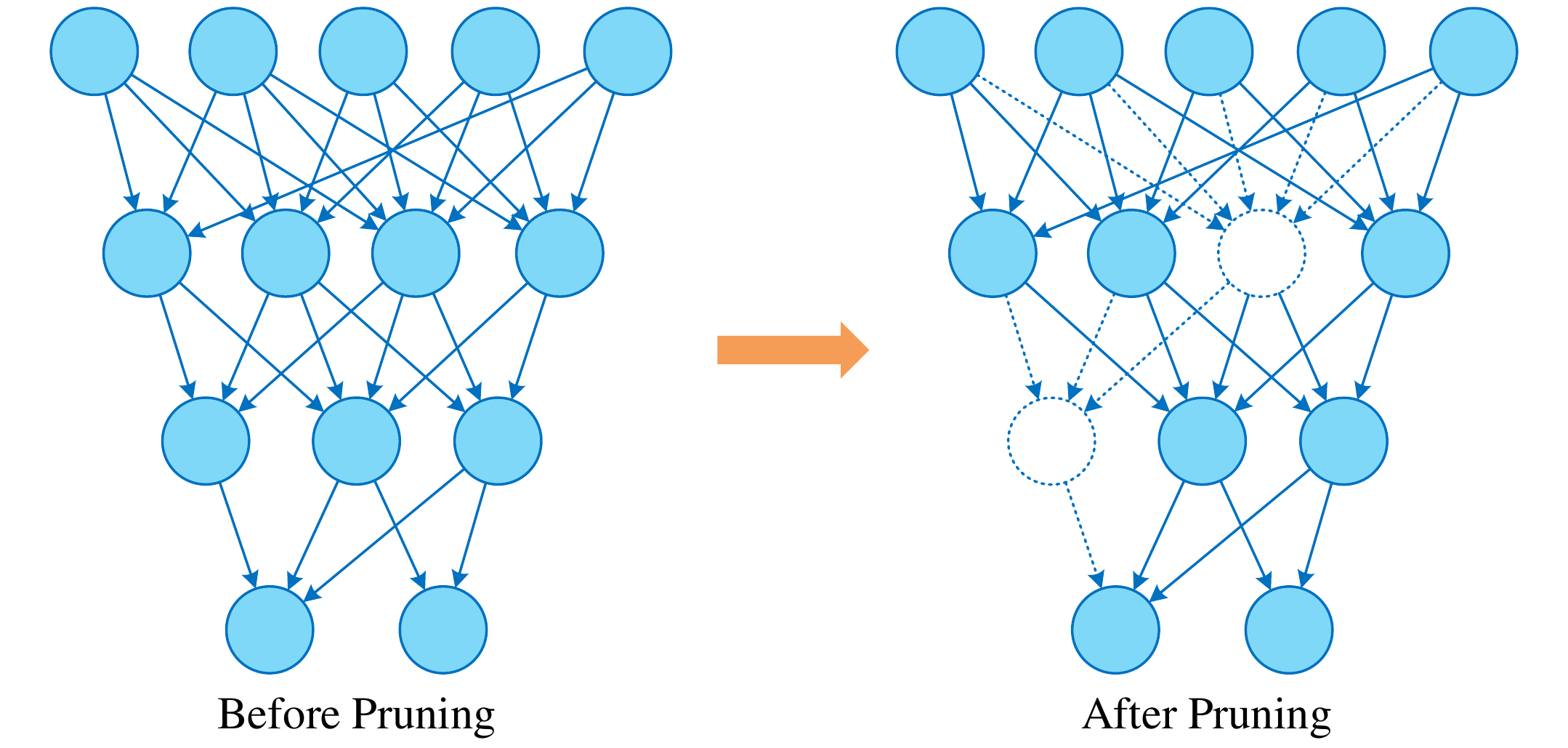} 
    \caption{Pruning connections and neurons. The dashed lines indicate the pruned parts.} 
    \label{fig3-2} 
\end{figure}

The magnitude-based pruning method makes the weights sparse by removing some unimportant connections or neurons, as shown in Fig.~\ref{fig3-2}. The pruning criterion in magnitude-based pruning greatly impacts final results. Meanwhile, special software and hardware may achieve better performance. In Table~\ref{tab:table3-4}, we briefly compare some magnitude-based pruning methods.

\begin{table}[htp]
    \centering
    \caption{Comparison of magnitude-based pruning methods.}
    \renewcommand{\arraystretch}{1.25}
    \label{tab:table3-4}
    \begin{tabular}{c|cccc}
    \hline
    \textbf{References}   & \begin{tabular}[c]{@{}c@{}}\textbf{Baseline} \\\textbf{models}\end{tabular}  & \textbf{Dataset}  & \textbf{FLOP}\% & \begin{tabular}[c]{@{}c@{}}\textbf{Compression} \\\textbf{rate}\end{tabular} \\ \hline
    \multirow{3}{*}{Han et al.~\cite{han_learning_2015}} & LeNet           & MNIST    & 16\%   & 12x              \\
                                               & AlexNet         & ImageNet & 30\%   & 9x               \\
                                               & VGG16           & ImageNet & 21\%   & 13x              \\ \hline
    \multirow{3}{*}{Han et al.~\cite{han_deep_2015}} & LeNet           & MNIST    & -      & 40x              \\
                                               & AlexNet         & ImageNet & -      & 35x              \\
                                               & VGG16           & ImageNet & -      & 49x              \\ \hline
    Guo et al.~\cite{guo_dynamic_2016}                  & AlexNet         & ImageNet & -      & 17.7x            \\ \hline
    \multirow{2}{*}{Li et al.~\cite{li_pruning_2016}}  & VGG16           & CIFAR10  & 34\%   & -                \\
                                               & ResNet          & CIFAR10  & 38.6\% & -                \\ \hline
    \end{tabular}
\end{table}

\begin{figure}[t] 
    \centering 
    \includegraphics[width=0.45\textwidth]{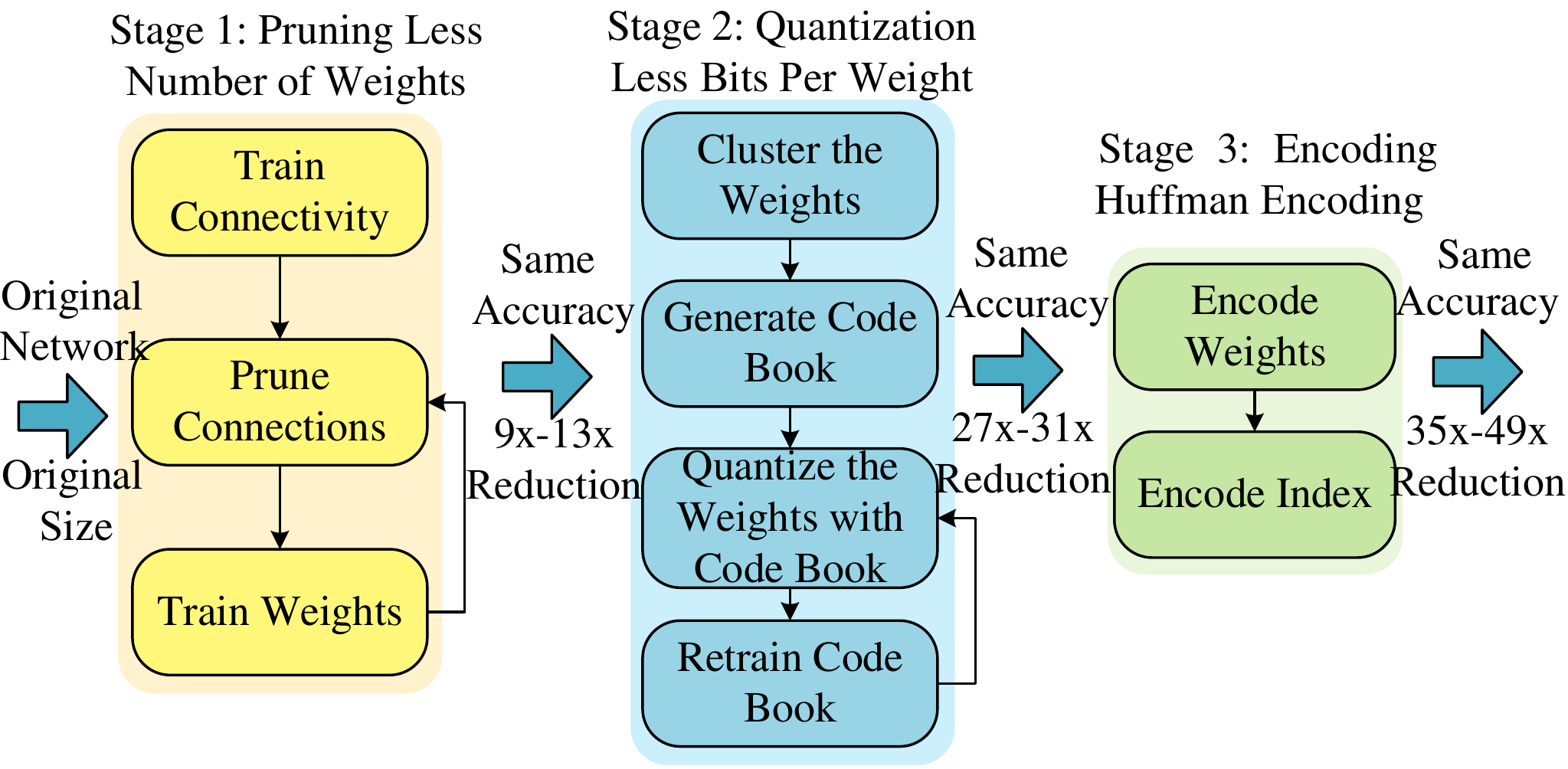} 
    \caption{The three stages compression pipeline (adapted from~\cite{han_deep_2015}). Compression is achieved through pruning, quantization, and encoding.} 
    \label{fig3-3} 
\end{figure}

Han et al.~\cite{han_learning_2015} proposed to measure the importance of different weights according to a preset threshold and prune the connections whose weights are below the threshold. Han et al.~\cite{han_deep_2015} proposed to combine pruning, quantization, and Huffman coding methods to get a more compact model. Fig.~\ref{fig3-3} summarizes the whole process. Based on~\cite{han_learning_2015}, Guo et al.~\cite{guo_dynamic_2016} pointed out that as the network changes, the importance of parameters will also change, so they proposed splicing to recover performance after deleting vital connections. Li et al.~\cite{li_pruning_2016} proposed to remove the convolution kernel that has little effect on the output accuracy and its connected feature maps without causing sparse connections. Hu et al.~\cite{hu_network_2016} defined an average percentage of zeros as a criterion for evaluating whether the convolution kernel is important. Yang et al.~\cite{yang_structured_2019} evaluated importance by introducing a mask for all convolution kernels. The mask's value decides the essentiality of every kernel. Lee et al.~\cite{lee_snip_2018} considered the absolute value of the derivative of the normalized objective function as a criterion for measuring the importance. Ref.~\cite{verma_network_2020} presented to use a multitasking network as a pruner to prune a pre-trained target network. This approach finishes the pruning procedure in one go instead of iterative pruning. 

\subsubsection{Channel Pruning}

Channel pruning is different from neuron-level pruning. Compared to removing a single neuron connection, pruning the entire channel has at least two advantages. First, it does not introduce sparsity, so no special software or hardware implementation is required for the generated model. Second, the inference phase does not require massive disk storage and runtime memory. Similar to magnitude-based pruning, the criterion used for channel pruning plays a quite crucial role in the final performance.

\begin{algorithm}[hp]
\small
  \caption{Calculating information gain.}
  \label{alg:algorithm2}
  \KwIn{ Deep separable convolution unit: $\langle L_i, D_i, P_i\rangle$; Dataset: $D$.}
  \KwOut{channel information gain $G$.}
  According to data set $D$, calculate the original information entropy of the $O_i$ layer denoted as $\mathsf{Entropy}_o$;
  
  \While{$A_i$ layer has $j$-th channel}{
  Set the eigenvalues of all positions of the $j$-th channel in $A_i$ layer to $0$ to obtain the new layer $A_i'$;
  
  Convolve $A_i'$ with the previous $1 \times 1$ channel convolution layer to get a new output $O_i'$;
  
  Calculate information entropy for this $O_i'$ denoted as $\mathsf{Entropy}_n$;
  
  Get the channel information gain $G_j$ by equation $G_j = \mathsf{Entropy}_n - \mathsf{Entropy}_o$;
  
  $j=j+1$;
  }
\end{algorithm}

He et al.~\cite{he_channel_2017} used a channel selection algorithm to prune every layer of the model and a least square reconstruction algorithm to rebuild the output of every pruned layer. Chen et al.~\cite{chen_exploiting_2019} proposed a framework combining channel pruning and low-rank decomposition. Bao et al.~\cite{bao_convolutional_2019} proposed a compression algorithm based on channel sparsity to prune channels. Hu et al.~\cite{hu_mutil_2019} contributed to performing channel selection (in the layer-by-layer way) to minimize the reconstruction error of the feature maps of the base model and the pruned model. Zhou et al.~\cite{zhou_a_2019} proposed to perform a layer-by-layer evaluation according to the standards of mixed statistics firstly, then erase channels as well as the corresponding kernels with negative scores, eventually fine-tune them with the help of KD. Zhang et al.~\cite{zhang_a_2019} introduced a pruning algorithm based on channel selection, which utilizes information gain and a quickly-recovering method to reconstruct model performance. This channel selection algorithm is summarized in Algorithm~\ref{alg:algorithm2}.

\subsubsection{Other types of pruning}

Yang et al.~\cite{yang_designing_2017} pointed out the previous methods may not reduce energy consumption. So they introduced an energy-aware pruning algorithm that conducted the pruning process according to the consumption of energy. O’Keeffe et al.~\cite{okeeffe_evaluating_2018} changed the whole pruning process by performing fine-tuning at regular intervals to maintain network performance and then continue pruning. The pruning technique proposed by Manessi et al.~\cite{manessi_automated_2018} allows pruning during the back-propagation stage of network training, which can perform end-to-end learning and significantly reduce training time. Huang et al.~\cite{huang_data-driven_2018} scaled the output of a specific structure with a factor and added sparse regularization. This optimization problem was solved by an improved random accelerated gradient method.

\subsection{Low-rank Decomposition}

The correlation expresses the structural information of the matrix. If there is a correlation between the rows, there is redundant information. The low-rank decomposition method in model compression reduces redundant information by measuring the matrix correlation. The basic idea of low-rank factorization is to replace the original large weight matrix with multiple small ones and remove redundant information. Matrix-factorization approaches include SVD, Canonical Polyadic (CP) decomposition, Tucker decomposition, and so on. As one of the most commonly used methods, SVD is defined as follows:
\begin{equation}
\footnotesize
\mathrm{X}=\left[\begin{array}{lll}
    \mathrm{u}_{1} \mathrm{u}_{2} & \dots \mathrm{u}_{\mathrm{r}}
    \end{array}\right]\left[\begin{array}{lll}
    \lambda_{1} & & \\
    & \dots & \\
    & & \lambda_{\mathrm{r}}
    \end{array}\right]\left[\begin{array}{c}
    \mathrm{v}_{1}^{\mathrm{T}} \\
    \vdots \\
    \mathrm{v}_{\mathrm{r}}^{\mathrm{T}}
\end{array}\right].\end{equation}

The low-rank decomposition of the matrix has played a significant role in model compression and acceleration of the CNN. Jaderberg et al.~\cite{jaderberg_speeding_2014} utilized cross-channel and convolution kernel redundancy to complete the acceleration of CNN by constructing kernels of rank $1$. This method is independent of architecture. Denton et al.~\cite{denton_exploiting_2014} used the linear structure existing in the convolution kernel to derive approximate values to reduce the calculation costs. Zhang et al.~\cite{zhang_efficient_2015} proposed to use non-linear elements instead of approximate linear convolution kernel or linear response to reduce the complexity of the convolution kernel. Tai et al.~\cite{tai_convolutional_2016} introduced a new low-rank tensor decomposition algorithm to eliminate redundancy in convolution kernels. The algorithm can find a precise global optimizer for decomposition, which is more effective than the iterative method. Kim et al.~\cite{kim_compression_2015} proposed an overall network compression scheme, which mainly includes three steps: 1) performing a rank selection of the variational Bayes matrix decomposition, 2) Tucker decomposition on the kernel tensor, 3) fine-tuning to compensate the cumulative loss of accuracy. Fan et al.~\cite{fan_cscc_2019} pointed out that the network pruning operation and the Rectified Linear Units (ReLU) activation function may generate many zero values during the training process and put forward an algorithm called convolution segmentation compression. Wiedemann et al.~\cite{wiedemann_compact_2020} proposed an efficient representation of a matrix with low entropy statistics, which helps to reduce the model size and execution complexity.

\subsection*{\bf Summary and Insights}

In this section, we review quantization, network pruning, and low-rank decomposition techniques. From the collected literature, we have the following observations.

1) \emph{The essence of quantization is to reduce the occupied space}. There are two methods, scalar and vector quantization, and fixed-point quantization. The former is that multiple weights share one weight, but the shared weight value needs to be restored to its original position during inference. That is, the decoding process is increased, which cannot save inference time. The latter uses fixed-point numbers to map floating-point numbers, which can save runtime memory and inference time. However, the former compresses the model itself to a much higher degree than the latter. It is more suitable for occasions where hardware storage space is scarce. The overall network structure is relatively simple, such as earphones and smartwatches. The latter is suitable for most occasions.

2) \emph{Network pruning attempts to increase the sparsity and achieves competitive accuracy compared to the baseline models}. This technique suits many pre-trained neural network models but still has many defects. It is usually hard to determine the criterion which determines the upper limit of the model's performance after pruning. Several existing studies introduce additional hyperparameters while the adjustment of hyperparameters is time-consuming. Meanwhile, repetitive fine-tuning is required for performance recovery after pruning, which is also time-consuming.

3) \emph{The low-rank decomposition technology can be effectively applied to the compression and acceleration of the fully connected layer}. However, there is currently no particularly effective implementation method for the convolutional layer. Low-rank decomposition is computationally expensive and not easy to implement. Furthermore, parameter compression can only be performed layer by layer.

\begin{figure*}[t] 
    \centering 
    \includegraphics[width=0.98\textwidth]{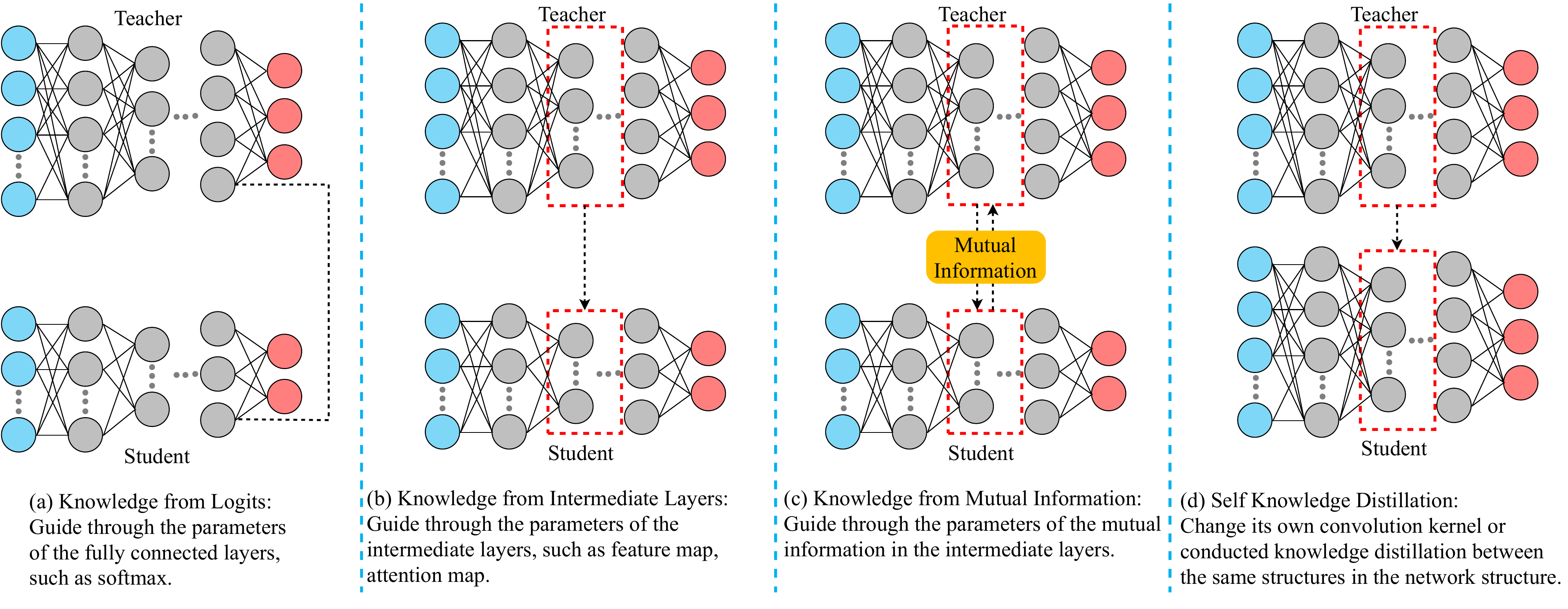} 
    \caption{Representative KD structures. The difference between four structures lies in the different sources of knowledge.} 
    \label{fig4-1} 
\end{figure*}

\section{Knowledge Distillation (KD)}
\label{sec:kd}

The basic idea of KD is to transfer the dark knowledge in the complex teacher model to the simple student model. Generally speaking, the teacher model has strong ability and performance, while the student model is compact. Through KD, the student model can have an approximated (or even surpassed) performance as the teacher model to achieve similar prediction results with less complexity. The general algorithm for KD training is shown in Algorithm~\ref{alg:algorithm3}. In recent years, diverse KD methods have been developed. 
We categorize major KD methods into the following types: \emph{A) knowledge from logits}, \emph{B) knowledge from intermediate layers}, \emph{C) mutual information distillation}, \emph{D) self-KD}, and \emph{E) other KD methods}. Fig.~\ref{fig4-1} depicts several representative KD structures.

\begin{algorithm}[t]
\footnotesize
  \caption{\small Algorithm for general KD training.
  The algorithm receives as input the trained parameters $\mathbf{W}_{\mathbf{T}}$ of a teacher, the randomly initialized parameters $\mathbf{W}_{\mathbf{S}}$ of a student, and two indices $h$ and $g$ corresponding to hint/guided layers, respectively. Let $\mathbf{W}_\mathbf{{Hint}}$ be the teacher’s parameters up to the hint layer $h$. Let $\mathbf{W}_\mathbf{{Guided}}$ be the student’s parameters up to the guided layer $g$.}
  \label{alg:algorithm3}
  \KwIn{$\mathbf{W}_{\mathbf{S}}$; $\mathbf{W}_{\mathbf{T}}$; $h$; $g$.}
  \KwOut{$\mathbf{W}_{\mathbf{S}}^{*}$.}
  
  $\mathbf{W}_{\mathbf{Hint}} \leftarrow\left\{\mathbf{W}_{\mathbf{T}}^{1}, \ldots, \mathbf{W}_{\mathbf{T}}^{h}\right\}$;
  
  $\mathbf{W}_{\mathbf{Guided}} \leftarrow\left\{\mathbf{W}_{\mathbf{S}}^{1}, \ldots, \mathbf{W}_{\mathbf{S}}^{g}\right\}$;
  
  $\mathbf{W}_{\mathbf{Guided}}^{*} \leftarrow\underset{\mathbf{W}_\mathbf{{Hint}}} {\operatorname{argmin}} \mathcal{L}\left(\mathbf{W}_{\mathbf{Guided}}\right)$;
  
  $\mathbf{W}_{\mathbf{S}}\leftarrow\left\{\mathbf{W}_{\mathbf{Guided}}^{* 1}, \ldots, \mathbf{W}_{\mathbf {Guided}}^{* g}\right\}$;
  
  $\mathbf{W}_{\mathbf{S}}^{*} \leftarrow\underset{\mathbf{W}_\mathbf{{S}}} {\operatorname{argmin}} \mathcal{L}\left(\mathbf{W}_{\mathbf{S}}\right)$.
\end{algorithm}

\subsection{Knowledge from Logits}

Fig.~\ref{fig4-1}\textcolor{red}{(a)} depicts the method of KD from logits. The neural network proposed by Hinton et al.~\cite{hinton_distilling_2015} usually uses softmax output layer to generate class probability. The output layer converts each class's logit, $z_i$, to probability $q_i$ by comparing it with other logits. In particular, we have,
\begin{equation} \small q_{i}=\frac{\exp \left(z_{i} / T\right)}{\sum_{j} \exp \left(z_{j} / T\right)},\end{equation} 
where $T$ is the distillation temperature and normally set to $1$. Using a higher value for $T$ produces a softer probability distribution on the class. Each case of transfer concentration contributes a cross entropy gradient $d c / d z_{i}$ to each logit and $z_i$ of the distillation model If the redundancy model has logits $v_i$ that generate the soft target probability $p_i$, and the transfer training is conducted at temperature $T$, the gradient is composed of:
\begin{equation}\small
\frac{\partial C}{\partial z_{i}}=\frac{1}{T}\left(q_{i}-p_{i}\right)=\frac{1}{T}\left(\frac{\exp \left(z_{i} / T\right)}{\sum_{j} \exp \left(z_{j} / T\right)}-\frac{\exp \left(v_{i} / T\right)}{\sum_{j} \exp \left(v_{j} / T\right)}\right).\end{equation} 
If the distillation temperature is higher than the logit, it can be approximated:
\begin{equation}\small \frac{\partial C}{\partial z_{i}} \approx \frac{1}{T}\left(\frac{1+z_{i} / T}{N+\sum_{j} z_{j} / T}-\frac{1+v_{i} / T}{N+\sum_{j} v_{j} / T}\right).\end{equation} If assumed that logits are zero for each transmission case, than $\Sigma_{j} z_{j}=\Sigma_{j} v_{j}=0$, 
the above formula is simplified as follows:
\begin{equation}\small \frac{\partial C}{\partial z_{i}} \approx \frac{1}{N T^{2}}\left(z_{i}-v_{i}\right).\end{equation}

Hinton et al.~\cite{hinton_distilling_2015} first proposed a KD framework based on knowledge transfer~\cite{cristian_model_2006}. They learned useful information from large models to train small models with close performance. This framework compresses a group of teacher networks into a student network with similar depth. Huang et al.~\cite{huang_like_2017} improved the neural selectivity transfer on the basis of knowledge dismantling, aiming at the pain point of KD, which is only applicable to the classification of softmax. Yu et al.~\cite{yu_learning_2019} proposed two new loss functions to simulate the communication between the deep teacher network and the small student network: one is based on the absolute teachers, the other is based on the relative teacher network. Mirzadeh et al.~\cite{mirzadeh_improved_2019} introduced multi-step knowledge extraction technology and used a medium-sized network (teacher assistant) to fill the gap between students and teachers.

\subsection{Knowledge from Intermediate Layers}

\begin{table*}[ht]
  \centering
  \caption{Comparison of KD based on knowledge of distillation sources and related details.}
  \renewcommand{\arraystretch}{1.25}
  \label{tab:table4-1}
  
  \begin{tabular}{c|cl}
  \hline
  \textbf{References}                      & \textbf{Knowledge from}         & \textbf{Details}                                                          \\ \hline
  Hinton et al.~\cite{hinton_distilling_2015}    & Logits                 & Cross Entropy                                                    \\ \hline
  Huang et al.~\cite{huang_like_2017}     & Logits                 & Cross Entropy and maximum mean discrepancy                                              \\ \hline
  Yu et al.~\cite{yu_learning_2019}        & Logits                 & Hints and attention                                              \\ \hline
  Mirzadeh et al.~\cite{mirzadeh_improved_2019}  & Logits                 & Use teacher assistant                                            \\ \hline
  Romero et al.~\cite{romero_fitnets_2015}    & Intermediate layers    & MSEloss in a certain middle layer                                \\ \hline
  Yim et al.~\cite{yim_a_2017}       & Intermediate layers    & Gram matrix loss in multiple middle layers                       \\ \hline
  Zagoruyko et al.~\cite{zagoruyko_paying_2017} & Intermediate layers    & Attention transfer loss in multiple middle layers                \\ \hline
  Zhang et al.~\cite{zhang_knowledge_2017}     & Intermediate layers    & Adaptive selection a middle layer                                \\ \hline
  Peng et al.~\cite{peng_correlation_2019}      & Mutual information     & Correlation between multiple instances                           \\ \hline
  Crowley et al.~\cite{crowley_moonshine_2019}  & Self structures        & With the same structure, use cheap convolution blocks            \\ \hline
  Park et al.~\cite{park_relational_2019}     & Structured knowledge   & Use a relational potential function to transfers the information \\ \hline
  Lopez-Paz et al.~\cite{vapnik_learning_2015}& Privileged information & Use pair-wise distillation  and holistic distillation            \\ \hline
  \end{tabular}
\end{table*}

This method registers directly between the teacher network and the student network by fitting features and feature maps. Fig.~\ref{fig4-1}\textcolor{red}{(b)} depicts the method of KD from intermediate layers. Romero et al.~\cite{romero_fitnets_2015} distilled a wide and deep network into a thin and deep network. The method proposed by Yim et al.~\cite{yim_a_2017} was not fitting the output of a large model but fitting the relationship between a large model and a small model layer by layer to refine the teacher network and the student network. The structure of the model proposed by Zagoruyko et al.~\cite{zagoruyko_paying_2017} allows the teacher network to guide the student network's attention maps learning by generating the attention maps. Zhang et al.~\cite{zhang_knowledge_2017} proposed a teaching KD method that is effective on small datasets. Shen et al.~\cite{shen_meal_2018} proposed using a confrontation-based learning strategy to extract different knowledge from different training models.

\subsection{Mutual Information Distillation}

The method of mutual information distillation is shown in Fig.~\ref{fig4-1}\textcolor{red}{(c)}. Peng et al.~\cite{peng_correlation_2019} proposed correlation congruence for KD, which transmits the information at the instance level and the correlation between instances. Ahn et al.~\cite{ahn_variational_2019} considered maximizing the lower bound of mutual information change between two neural networks and proposed an information theory framework for knowledge transfer. Tung et al.~\cite{tung_similary_2019} proposed a new form of KD loss, which was inspired by a similar input pattern in a well-trained network.

\subsection{Self Distillation}

The method of self distillation is shown in Fig.~\ref{fig4-1}\textcolor{red}{(d)}. The training of the self-distillation framework directly points to the student model. Therefore, it not only provides less training time but also has higher accuracy. Crowley et al.~\cite{crowley_moonshine_2019} replaced the original convolution blocks with cheap convolution blocks while keeping the same student network as teacher network. The authors showed that with the same parameters, the student network had better results than the teacher network. Zhang et al.~\cite{zhang_be_2019} proposed a self-distillation training framework to improve the accuracy of the model.

\subsection{Other KD methods}

The Graph-based distillation method is a KD technique used in graph neural networks. Lee et al.~\cite{lee_graph-based_2019} proposed a new method for extracting dataset-based knowledge from teacher networks using attention networks. Ma et al.~\cite{ma_graph_2019} proposed a multi-task knowledge extraction method for graph representation learning that uses graph metrics based on network theory as an auxiliary task through multi-task learning.

Meanwhile, Park et al.~\cite{park_relational_2019} proposed distance and angular distillation losses to compensate for structural differences in the relationship. Tian et al.~\cite{tian_contrastive_2019} captured correlation and high-order output correlation by comparing target families and adapted them to extract knowledge from one neural network to another. Liu et al.~\cite{liu_structured_2019} considered transferring structural information from large networks to small networks for intensive prediction tasks and proposed paired distillation with paired similarities by establishing static diagrams and global distillation with antagonistic training to extract overall knowledge. Gao et al.~\cite{gao_residual_2020} proposed residual KD, which further distilled the knowledge by introducing an assistant.

Moreover, Vapnik et al.~\cite{vapnik_learning_2015} considered a learning paradigm called learning using privileged information. Lopez-Paz et al.~\cite{lopez-paz_unifying_2015} unified distillation and privileged information as generalized distillation, i.e., a framework for learning data representation from multiple machines. Tang et al.~\cite{tang_retaining_2019} proposed a method for multitask learning to preserve privileged information.

Wang et al.~\cite{Wang_kdgan_2018} and Xu et al.~\cite{xu_training_2018} combine KD with GAN. Uijlings et al.~\cite{uijlings_revisiting_2018} proposed that a group of source classes with boundary box annotations be used to revisit the knowledge transfer of training object detectors on the target class of weakly supervised training images. Anil et al.~\cite{anil_large_2018} proposed large-scale distributed neural network training through online distillation. Furlanello et al.~\cite{furlanello_born_2018} trained and parameterized the same student network as the teacher network, making the student network outperform the teacher network. Tan et al.~\cite{tan_learning_2018} used model distillation to learn a global additive explanation describing the relationship between input characteristics and model predictions. Heo et al.~\cite{heo_knowledge_2018} proposed a method for knowledge transfer by extracting activation boundaries formed by cryptogenic neurons. He et al.~\cite{he_knowledge_2019} made some improvements to traditional distillation methods for the semantics split task. Liu et al.~\cite{liu_knowledge_2019} transferred ``knowledge'' from multiple deep teacher networks to a deep network student network. Yang et al.~\cite{yang_knowledge_2020} proposed a KD method based on transferring feature statistics from the teacher network to the student network. Song et al.~\cite{song_lightpaff_2020} proposed a framework that transfers knowledge from a large pre-training model to a small model and performs both pre-training and fine-tuning at the same time.

\begin{figure}[t] 
    \centering 
    \includegraphics[width=0.48\textwidth]{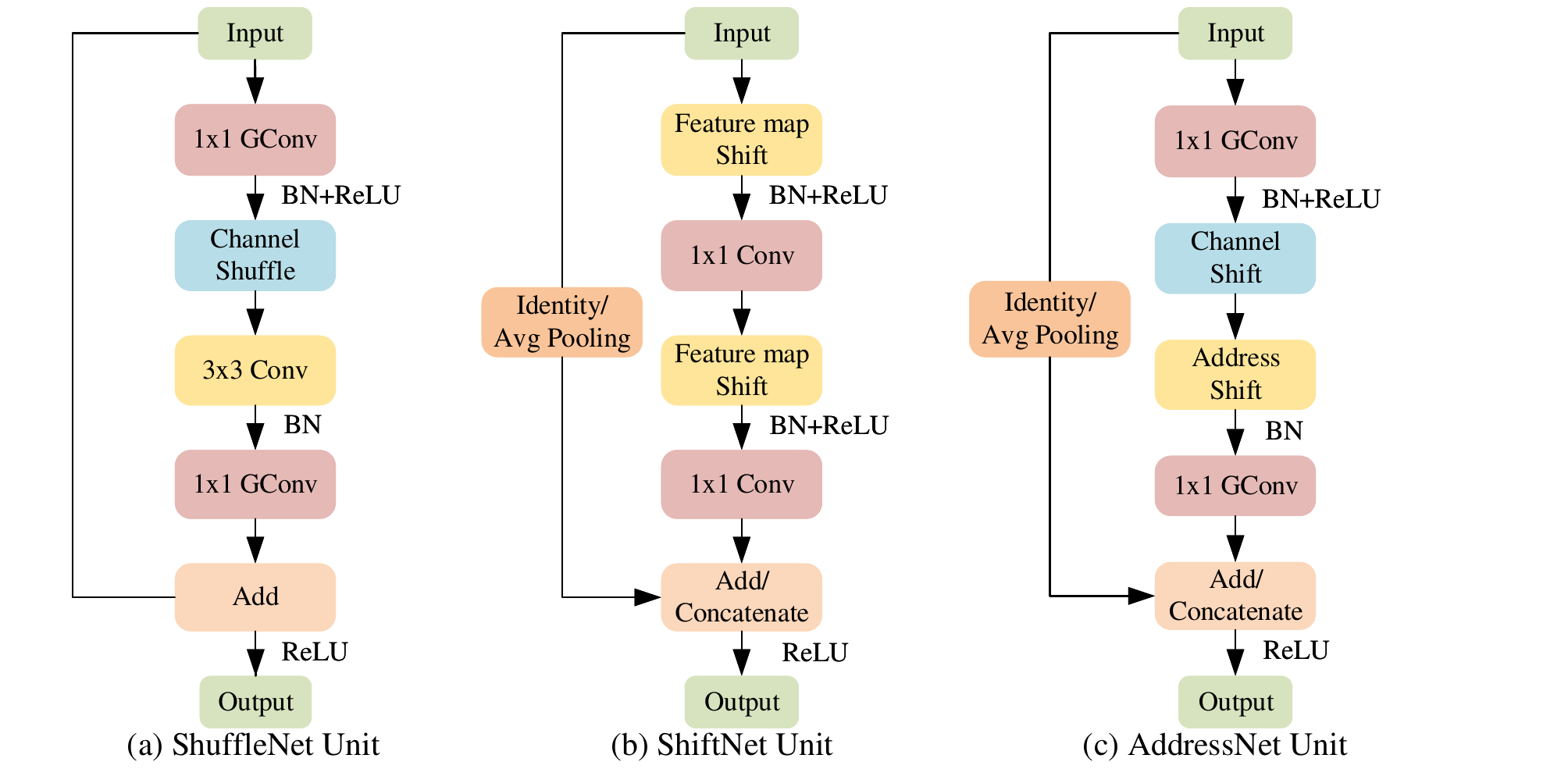} 
    \caption{Basic units of ShuffleNet, ShiftNet and AddressNet (adapted from~\cite{he_addressnet_2019}), they replaced the $1 \times 1$ convolution, $3 \times 3$ convolution, and $1 \times 1$ and $3 \times 3$ convolution units based on the bottleneck structure, respectively.} 
    \label{fig5-1} 
\end{figure}

\subsection*{\bf Summary and Insights}

In this section, we review the most representative KD methods. Table~\ref{tab:table4-1} compares several representative KD methods from knowledge extraction and implementation details. The fully-connected layer plays a role of ``extracting knowledge from logits'' while softmax is usually used. In general, KD methods are more suitable for classification networks. ``Knowledge from intermediate layers'' makes full use of the information in the teacher network to better guide the student network. 

Compared with the methods of obtaining knowledge from logits, obtaining knowledge from the intermediate layers is more general. ``Mutual information'' is an improvement based on ``intermediate layers''. Mutual information can strengthen the interdependence between the teacher network and the student network, thereby reducing training time. In ``Self distillation'', the structure of the teacher and student network is almost the same. The student network can either replace the teacher network's convolution blocks with cheap ones or adopt the teacher's shallow feature maps to guide the whole knowledge distillation process.

The essence of KD lies in the fact that the student network uses soft tags provided by the teacher network for better training. Therefore, both the teacher network's structure and the mechanism of information using have a significant impact on the KD methods' effect. Only a suitable network and a suitable KD method can cooperatively lead to outstanding performance.

\section{Modification of network structures}
\label{sec:modif}

In this section, we explore how to design lightweight models by modifying the network structure. This process can be achieved through modifications of channels, filters, connections between the neurons, the active functions, and other components. Table~\ref{tab:table5-1} compare the representative network structures.

\subsection{Channel Shuffle/Shift}

The compact DNNs can be achieved by modifying the network structure through \emph{channel shuffle} or \emph{shift}. Zhang et al.~\cite{zhang_shufflenet_2018} designed a novel compact network architecture for mobile devices called ShuffleNet, as shown in Fig.~\ref{fig5-1}\textcolor{red}{(a)}. Since feature maps are more important than $1 \times 1$ convolutions for small models, they applied a group convolution mechanism to the $1 \times 1$ pointwise convolutions to reduce the computational cost required for $1 \times 1$ convolutions. To avoid the group convolution's boundary effect, the authors came up with a new group convolution being followed by a channel shuffle operation to help to effectively mix each group's information flow in the group convolution.

In contrast, Wu et al.~\cite{wu_shift_2018} introduced ShiftNet, as shown in Fig.~\ref{fig5-1}\textcolor{red}{(b)}. In this design, the spatial convolutions extract the space information in different channels and the kernel size has a great influence on the model's size and computation cost. As a result, they proposed to shift the feature map in different directions as an alternative to spatial convolution. In this way, they could achieve a similar spatial convolution function in a parameters-free and FLOP-free channels way and got good performance in lightweight models. He et al.~\cite{he_addressnet_2019} absorbed benefits from the previous two studies and proposed AddressNet. The authors found that a simple reduction of the FLoating-point Operations Per Second (FLOPS) and parameters does not shorten the inference time. Therefore, they adopted three time-saving shift operations as the alternative of channel shuffle, spatial convolution, and shortcut connection to alleviate the bottleneck. Fig.~\ref{fig5-1}\textcolor{red}{(c)} shows the basic unit of AddressNet, which replaces the channel shuffle operation with the channel shift operation. 

\begin{figure}[t] 
    \centering 
    \includegraphics[width=0.48\textwidth]{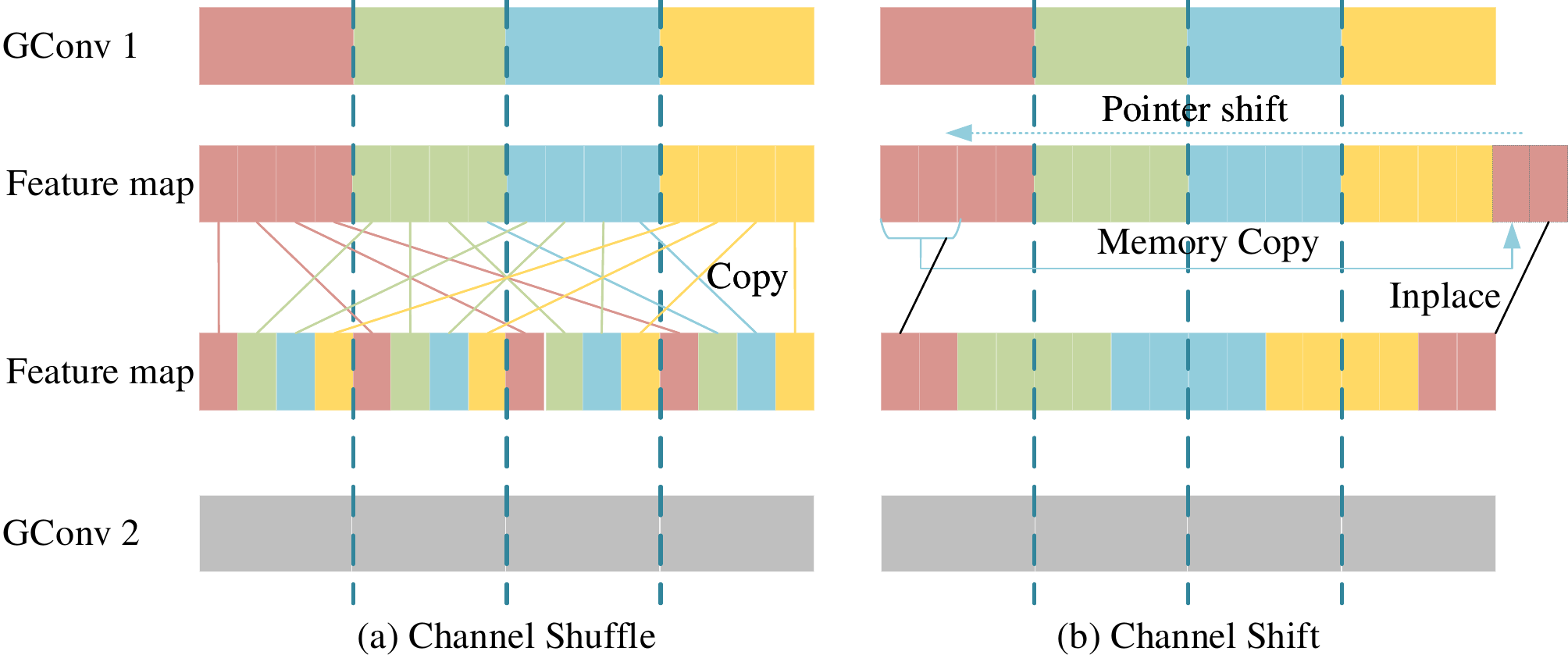} 
    \caption{The compare between Channel Shuffle and Channel Shift (adapted from~\cite{wu_shift_2018}). Channel shuffle needs to copy and transfer all feature map data, while channel shift only needs to move the data of two units and the offset of the start pointer by two units.} 
    \label{fig5-2} 
\end{figure}

Fig.~\ref{fig5-2} shows the difference between these two operations. Although the two processes are not the same, the experimental results prove that they have a similar accuracy. Meanwhile, the latter is more time-saving than the former. Secondly, it replaces the $3 \times 3$ convolution by address shift, which is identical to the feature map shift operation of ShiftNet while only having four directions. Finally, based on the shortcut of the bottleneck module, they proposed the shortcut shift, allocating a piece of continuous space in advance for the concatenations operation to further reduce the time overhead.

\subsection{Shortcut Connections}

Shortcut connections, also called residual connections, were proposed in~\cite{he_deep_2016} to solve the problem of vanishing gradients in DNNs. Those connections can integrate lower-dimensional and more detailed information with higher-dimensional information and provide a shorter path for back-propagation, making the error propagation smoother.

He et al.~\cite{he_deep_2016} found that the problem of vanishing gradients heavily affects the network convergence and the network degradation also becomes heavier when the network goes deeper. To solve the above problems, they proposed the deep residual learning framework and introduced residual mapping. Assume that the original mapping is $H(x)$. They introduce a residual mapping: $F(x)=H(x)-x$, and then add $x$ to the output of $F(x)$ with the shortcut connection. As a result, the original mapping $H(x)$ is transformed into $F(x)+x$. It made the training process easier. Due to the shortcut connection, the shallower layers can directly receive the error from the deeper layers, without worrying about the situation that the error gradient tends to $0$ as the depth increases. The entire process does not introduce additional parameters and computational complexity. In 2017, an improved version called ResNeXt~\cite{xie_aggregated_2017} was proposed to ameliorate the topology design to obtain higher accuracy with less complexity.

Since ResNet and some other models demonstrated the benefit from shortcut connections, Huang et al.~\cite{huang_densely_2017} proposed DenseNet based on a similar idea. In DenseNet, the input of each layer is a concatenation of the outputs of all preceding layers, consequently obtaining more abundant information for the input to achieve better performance. since there are some redundant connections in DenseNet, Huang et al.~\cite{huang_condensenet_2018} proposed CondenseNet by combining DenseNet with learned group convolutions. The learned group convolutions find the unimportant input connections in DenseNet and then prune them. Zhao et al.~\cite{zhao_interpretable_2019} proposed a rule for customizing the number of feature maps, introducing a compression layer composed of $1 \times 1$ convolution, and combining learned group convolutions to achieve parameter reduction. Moreover, Zhu et al.~\cite{zhu_sparsely_2019} proposed SparseNet, in which only $\log(l)$ previous output feature maps would concatenate as the $l$ layer's input at any given location in the network. As a result, the number of connections is further reduced compared to DenseNet.

\begin{table*}[ht]
    \centering
    \caption{Comparison of Different Structures.}
    \renewcommand{\arraystretch}{1.25}
    \label{tab:table5-1}
    \footnotesize
    \begin{tabular}{c|c|ccc}
    \hline
    \textbf{Models} &
      \textbf{Basic structures} &  \textbf{Improved components} &  \textbf{Improved methods} &  \textbf{Reduction/Improvement} \\ \hline
    \begin{tabular}[c]{@{}c@{}}Zhang et al.~\cite{zhang_shufflenet_2018} \\ ShuffleNet\end{tabular} &
      \multirow{2}{*}{\begin{tabular}[c]{@{}c@{}}Depth-wise \\ separable \\ convolutions\end{tabular}} &
      \begin{tabular}[c]{@{}c@{}}$1 \times 1$ \\ convolutions\end{tabular} &
      \begin{tabular}[c]{@{}c@{}}Group convolution, \\ channel shuffle, \\ shortcut connections\end{tabular} &
      Wider feature maps \\ \cline{1-1} \cline{3-5} 
    \begin{tabular}[c]{@{}c@{}}Wu et al.~\cite{wu_shift_2018} \\ ShiftNet\end{tabular} &
       &
      \begin{tabular}[c]{@{}c@{}}$3 \times 3$ \\ convolutions\end{tabular} &
      \begin{tabular}[c]{@{}c@{}}Feature map shift, \\ shortcut connections\end{tabular} &
      Parameters and computational cost \\ \hline
    \begin{tabular}[c]{@{}c@{}}He et al.~\cite{he_addressnet_2019} \\ AddressNet\end{tabular} &
      \begin{tabular}[c]{@{}c@{}}ShuffleNet \\ and \\ ShiftNet\end{tabular} &
      \begin{tabular}[c]{@{}c@{}}$3 \times 3$ \\ convolutions \\ and feature \\ map shift\end{tabular} &
      \begin{tabular}[c]{@{}c@{}}Address shift, \\ channel shift, \\ short shift\end{tabular} &
      Inference time \\ \hline
    \begin{tabular}[c]{@{}c@{}}He et al.~\cite{he_deep_2016} \\ ResNet\end{tabular} &
      \multirow{5}{*}{\begin{tabular}[c]{@{}c@{}} \\ \\ \\ \\ \\ \\  Conventional \\ convolution\end{tabular}} &
      \begin{tabular}[c]{@{}c@{}}Mapping \\ rules\end{tabular} &
      \begin{tabular}[c]{@{}c@{}}Shortcut connections/\\ bottleneck structure\end{tabular} &
      Training time, deeper network \\ \cline{1-1} \cline{3-5} 
    \begin{tabular}[c]{@{}c@{}}Huang et al.~\cite{huang_densely_2017} \\ DenseNet\end{tabular} &
       &
      \begin{tabular}[c]{@{}c@{}}Connections, \\ kernels, \\ input integration\end{tabular} &
      \begin{tabular}[c]{@{}c@{}}Fully-connected, \\ narrower kernels, \\ add to concatenation\end{tabular} &
      Parameters and feature reuse \\ \cline{1-1} \cline{3-5} 
    \begin{tabular}[c]{@{}c@{}}Hu et al.~\cite{hu_squeeze_2018} \\ SENet\end{tabular} &
       &
      \begin{tabular}[c]{@{}c@{}}Channel \\ dependencies\end{tabular} &
      \begin{tabular}[c]{@{}c@{}}Training channels \\ selectively\end{tabular} &
      Representation capability \\ \cline{1-1} \cline{3-5} 
    \begin{tabular}[c]{@{}c@{}}Iandola et al.~\cite{iandola_squeezenet_2016} \\ SqueezeNet\end{tabular} &
       &
      \begin{tabular}[c]{@{}c@{}}Convolutional \\ rules\end{tabular} &
      \begin{tabular}[c]{@{}c@{}}Reformulated kernels \\ distribution \\ (reduce $3 \times 3$ kernels) \\ and stride\end{tabular} &
      Parameters and bigger Activation Maps \\ \cline{1-1} \cline{3-5} 
    \begin{tabular}[c]{@{}c@{}}Li et al.~\cite{li_sep-nets_2017} \\ SEPNets\end{tabular} &
       &
      \begin{tabular}[c]{@{}c@{}}Kernel \\ quantization\end{tabular} &
      \begin{tabular}[c]{@{}c@{}}Binarized $k \times k$ kernels, \\ shortcut connections\end{tabular} &
      Parameters and computational cost \\ \hline
    \begin{tabular}[c]{@{}c@{}}Chollet.~\cite{chollet_xception_2017} \\ Xception /\\ Howard et al.~\cite{howard_mobilenets_2017} \\ MobileNet\end{tabular} &
      \begin{tabular}[c]{@{}c@{}}Inception \\ family \\ models\end{tabular} &
      \begin{tabular}[c]{@{}c@{}}Convolutional \\ rules\end{tabular} &
      \begin{tabular}[c]{@{}c@{}}Depth-wise Separable \\ convolutions\end{tabular} &
      Parameters and computational cost \\ \hline
    \begin{tabular}[c]{@{}c@{}}Huang et al.~\cite{huang_condensenet_2018} \\ CondenseNet\end{tabular} &
      DenseNet &
      \begin{tabular}[c]{@{}c@{}}Full \\ connection\end{tabular} &
      \begin{tabular}[c]{@{}c@{}}Learned group \\ convolution\end{tabular} &
      Parameters \\ \hline
    \begin{tabular}[c]{@{}c@{}}Zhu et al.~\cite{zhu_sparsely_2019} \\ SparseNet\end{tabular} &
      \multirow{2}{*}{\begin{tabular}[c]{@{}c@{}} \\ \\ CondenseNet\end{tabular}} &
      \begin{tabular}[c]{@{}c@{}}Full \\ connection\end{tabular} &
      \begin{tabular}[c]{@{}c@{}}$\log(l)$ \\ connections\end{tabular} &
      Parameters \\ \cline{1-1} \cline{3-5} 
    \begin{tabular}[c]{@{}c@{}}Zhao et al.~\cite{zhao_interpretable_2019} \\ RSNet\end{tabular} &
       &
      \begin{tabular}[c]{@{}c@{}}Output \\ channel \\ number\end{tabular} &
      \begin{tabular}[c]{@{}c@{}}Adaptive output \\ channels under \\ the number of \\ input channels\end{tabular} &
      Parameters and features \\ \hline
    \begin{tabular}[c]{@{}c@{}}Wu et al.~\cite{wu_deep_2019} \\ DESNet\end{tabular} &
      SENet &
      \begin{tabular}[c]{@{}c@{}}Generalization \\ for multi-connected \\ structure\end{tabular} &
      \begin{tabular}[c]{@{}c@{}}Fully-connected \\ SENet\end{tabular} &
      Parameter usage \\ \hline
    \end{tabular}
\end{table*}

Coincidentally, Fooladgar and Fahimeh also proposed a lightweight structure named RDenseNet~\cite{fooladgar_lightweight_2020} after investigating ResNet and DenseNet. RDenseNet achieves impressive results on experiments on multiple datasets.

\subsection{Basic Network Units}
\subsubsection{Depth-wise Separable Convolutions}

In traditional convolution networks, convolutions need to complete the dual mapping of spatial correlation and channel correlation. Since 2014, Google gradually released Inception family models to explore ways to separate these two mapping correlations. And the Xception~\cite{chollet_xception_2017} is the extreme version for this design, where the channel correlation mapping and the spatial correlation mapping are completely separated. Moreover, the efficient depth-wise separable convolution unit was developed. Experiments showed that Xception achieves a higher utilization rate of the parameters, better performance, and accuracy than its predecessor - InceptionV3 with the same parameters.



Howard et al.~\cite{howard_mobilenets_2017} proposed the MobileNet and elaborated the compression capability of depthwise separable convolutions in terms of the parameters and computation cost. To improve their model's generalization ability, they also proposed two global parameters to adjust the width and calculation complexity to achieve a better trade-off between efficiency and accuracy in different application scenarios.

\begin{figure}[t] 
    \centering 
    \includegraphics[width=0.48\textwidth]{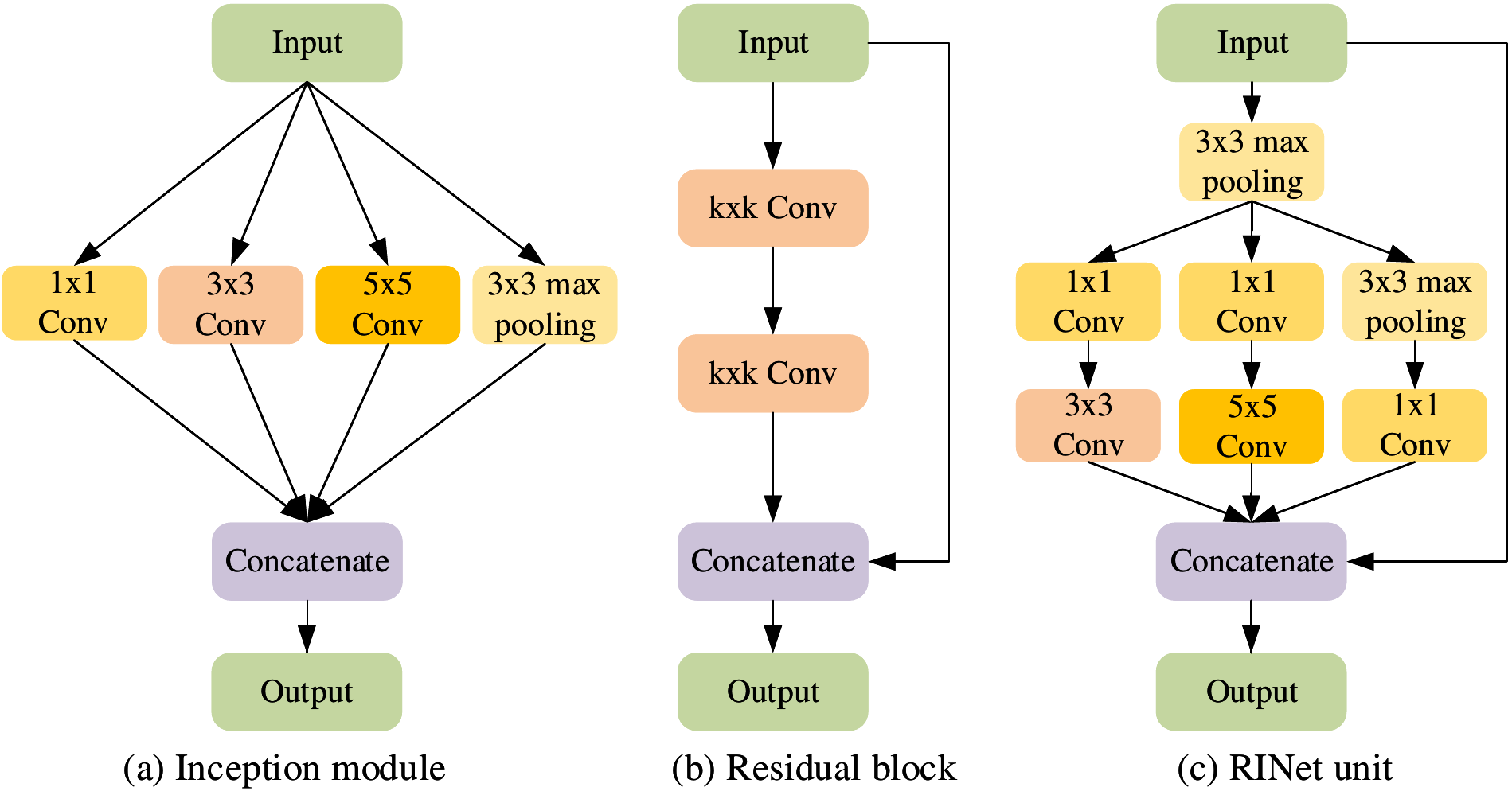} 
    \caption{The structures of Inception module, Residual block, and RINet unit (adapted from~\cite{gao_a_2019}), where he RINet unit is a combination of Inception module and Residual block.} 
    \label{fig5-3} 
\end{figure}

After studying the Inception series models and ResNet, Gao et al.~\cite{gao_a_2019} proposed RINet, a combination of both Inception and ResNet models. RINet adopts Inception's multi-branch multi-scale convolution structure and incorporates ResNet's residual connection structure to achieve computational reduction and training acceleration, as shown in Fig.~\ref{fig5-3}.

\subsubsection{SE Module}

Unlike most of the studies to improve the information representation capabilities of CNNs in terms of spatial correlation, Hu et al.~\cite{hu_squeeze_2018} chose to achieve this goal by starting from channel correlation. They released a new neural network called SENet, in which Squeeze-and-Excitation (SE) module is the basic unit. The SE module improves the network's ability of information representation by explicitly modeling the dependencies between channels and adaptively calibrating channels' feature responses. In this way, the essential features can be strengthened, consequently suppressing the unimportant features. Later, Wu et al.~\cite{wu_lightweight_2018} combined SENet and depthwise separable convolutional networks to propose an impressive network CT-SECN for super-resolution image processing.


Since SENet was designed mainly based on conventional CNNs, it is not suitable for the case that multiple network layers are connected each other and the feature maps are dynamically adjusted like ResNet and DenseNet. To address this issue, Wu et al.~\cite{wu_deep_2019} proposed a novel network called DESNet, which combines the advantages of DenseNet and SENet and demonstrates superior performance.


\subsubsection{Fire Module}

For the existing neural network models, it is easy to find multiple CNN models meeting the condition when an accuracy level is given. For the same accuracy level, smaller models usually have more flexibility, convenience, and low cost during the training and deploying process. To this end, Han et al.~\cite{iandola_squeezenet_2016} tried to design models with comparable accuracy but fewer parameters than the well-known models. Furthermore, they successfully trained a network referred to as SqueezeNet with the same precision as AlexNet while having $50\times$ fewer parameters. Based on SqueezeNet, Wu et al.~\cite{wu_squeezedet_2017} proposed a small and efficient convolution network called SqueezeDet for the auto-driving application.


\subsubsection{Other Modules}

In~\cite{li_sep-nets_2017}, Li et al. found that $k \times k$ kernels mainly extract features in a specific spatial pattern while they are less important than other components in the entire model. Therefore, they performed binarization on the $k \times k$ convolution kernels to cut down parameters. Meanwhile,~\cite{zhang_scan_2019} introduced scalable neural networks, which achieve neural network compression and acceleration simultaneously. Moreover, Li et al.~\cite{li_iirnet_2019} designed an intensely-inverted residual block unit, which introduces inverted residual structure and multi-scale low-redundancy convolution kernels. The inverted residual structure expands the input dimension at the first $1 \times 1$ layer so that the following depthwise layer can extract more features while the $1 \times 1$ group convolution layer compresses the features and reduces the computational complexity. In addition, Liu and Di~\cite{liu_tanhexp_2020} also proposed a new activation function $f(x)=x \tanh \left(e^{x}\right)$, which can achieve rapid convergence of the model, thereby improving the performance of image recognition.


\subsection{Network Architecture Search (NAS)}

NAS is a method to find the optimal network structure configuration in the search space through reinforcement learning. However, when the search space is enormous, the direct application of NAS results in a high computational cost. In general, neural networks often have the repetition of unit structures, e.g., the same convolution kernel group, the same non-linear unit structure, or even the same connection composition. Therefore, \emph{can we find a universal expression of a convolution unit through NAS and then stack the convolutional units together to achieve a high accuracy after fine-tuning them?}

Driven by this idea, Zoph et al.~\cite{zoph_learning_2018} proposed a network construction method that searches the basic unit of the network through NAS on a small dataset and then transfers this unit to a large dataset. They validated this method on the CIFAR10 and ImageNet datasets and obtained a new small network called NASNet. This approach not only reduces the time required for directly searching the optimal model on large datasets but also enhances the network's generalization ability because small units usually have stronger generalization capabilities. In~\cite{tan_efficientnet_2019}, Tan et al. studied the model-scaling problem and found that carefully-balancing the width, depth, and resolution of the network can effectively improve the performance of the network. They then design a primary network to achieve the balance of the three factors via the realization by NAS. To further reduce the search space, they proposed to use a composite parameter to uniformly scale these three dimensions.

\subsection*{\bf Summary and Insights}

This section reviews the commonly-used structures, operations, and modules in designing lightweight networks. We also introduce methods for finding a model's basic units from small datasets. Table~\ref{tab:table5-1} compares the representative network structures. Among the above-mentioned basic structures, shortcut connections and depthwise separable convolutional modules have been widely adopted with many derivative extensions. Meanwhile, both SE module and Fire module have also performed well. As an emerging lightweight network design tool, NAS has also been in the limelight and it is expected to receive a growing popularity in the future. The above methods that have been widely used in various network structures are orthogonal to other compacting-DNNs techniques. Thus, researchers shall integrate them together to achieve outstanding performance.

\section{Applications of Compacted DNNs in IoT}
\label{sec:app}

In this section, we describe application scenarios of compacting-DNNs technologies in IoT. We categorize those applications into basic services and upper applications.


\subsection{Basic Services}

Basic services can be categorized into the following types.

\subsubsection{Image Processing and Computer Vision (CV)}

There is a large amount of data in the form of images and videos in IoT applications. With the continuous development of mobile cameras, high-resolution images and videos have continued to be produced. Meanwhile, the advances of image sensors (inside a camera) also proliferate the wide adoption of diverse cameras in IoT scenarios, such as surveillance in cities, face recognition, and product-quality control (e.g., detecting flaws of products~\cite{9266587}). There are individual needs for different applications of collected images of cameras. 

In general, DL algorithms can be widely used for image processing and computer vision. DL methods can be used to improve images in aspects such as denoising, defogging, deblurring, enhancement, super-resolution, repairing, restoration, and coloring. Besides image-processing tasks, DL techniques can also be used in computer-vision tasks, such as semantic segmentation, instance segmentation, target detection, target recognition, and target tracking. Moreover, DL also plays an important role in video processing and augmented reality.

In the above applications, portable or compacted DL models are expected to be deployed at IoT devices or at mobile cameras. Take the identity-recognition application as an example~\cite{SAJJAD2020995}. There are wide deployments of face-recognition devices at public agencies for identity recognition, authentication, and authorization. However, it is challenging to directly deploy conventional DL-based face recognition models at resource-limited devices. Thus, compacted DNNs and portable DL models are a necessity to address this issue.


\subsubsection{Voice and Audio Processing}

We have also experienced the proliferation of many audio-processing applications in IoT. For example, various IoT devices deployed in a production line can be used to collect ambience sound, which can be used to identify possible flaws of the production line~\cite{KZhang:TII20,HuangTII2020Blockchain}. Meanwhile, human-computer interaction (HCI) applications also demonstrate the efficiency in interacting through voice, especially for voice recognition and voice process at mobile devices and wearable devices. DL methods also have the strengths in processing voice and audio information. 

In the above scenarios, it is necessary to deploy portable DL models at small devices, which have limited resources. Price et al.~\cite{price_14.4_2017} proposed a speech recognizer with DNN-based acoustic model. The specially designed hardware can more effectively store the sparse weight matrix. Moreover, the quantization is also exploited to reduce the consumption of memory. In addition, the speech recognition processor proposed by Zheng et al.~\cite{zheng_an_2019} adopts the idea of BCNN, in which the authors accelerated the processing of BCNN from the hardware level. The main approaches include 1) maximizing the reuse of data stream to minimize memory access, 2) pruning the bit-level regularization to compress the weight matrix. 3) using self-learning on the chip to update the weights at runtime.


\subsubsection{Indoor Localization}

The increasing demands for context-aware IoT applications lead to the flourishing indoor-localization services. Different from Global Positioning System (GPS), which heavily relies on satellite signals, indoor-localization services often are achieved by processing and analyzing radio signals, such as WiFi and Bluetooth at IoT devices since GPS signals cannot be well received indoor environment. It is also challenging to process indoor-localization radio signals at IoT devices. DL methods have also demonstrated their strengths in indoor-localization services.

Shao et al.~\cite{shao_indoor_2018} use a lightweight CNN model to achieve precise non-directional positioning. They mainly use WiFi signals and magnetic field fingerprints for indoor localization. After fusing the types of signals, they use a CNN for a localization task (i.e., a classification task). However, the training dataset is quite limited while there are a large number of classification points in this scene. To address this issue the authors also proposed a two-part training method. Meanwhile, Jiang et al.~\cite{jiang_deep_2019} used RFID and DBN for indoor positioning. The authors designed a DBN with four hidden layers for extracting features to estimate the position. Moreover, Samadani et al.~\cite{samadani_indoor_2020} used channel status information and DL methods for indoor positioning. 

\subsubsection{Physiological Monitoring}

The wide adoption of various wearable IoT devices and video cameras also boosts diverse physiological-monitoring applications, from entertainment, education, medical care, and industry domains. The physiological-monitoring applications often require human posture estimation and activity recognition. DL models can well process these physiological-monitoring tasks.

Tao et al.~\cite{tao_multicolum_2016} recognized human activities with the help of mobile devices. The authors proposed a bidirectional LSTM to obtain excellent results. Meanwhile, Zhu et al.~\cite{zhu_a_2019} used the data of inertial sensors in mobile devices to recognize human activities. The authors proposed a deep LSTM and adopted a semi-supervised learning method. 

Human activities often carry information with personal characteristics, which can be further used for identity verification. Deb et al.~\cite{deb_actions_2019} proposed a way of passive authentication in smartphones. When people use mobile phones, the data collected by the motion sensors in the mobile phones are passively authenticated with the help of the Siamese LSTM proposed by the authors. Moreover, Qin et al.~\cite{qin_a_2019} utilized human biometric gait information and DNN for authentication.

\subsubsection{Security and Privacy}

The wide application of DL models also poses security and privacy issues. There is a line of studies in providing privacy and security protection in different IoT scenarios. He et al.~\cite{he2017real} found that smart grid is vulnerable to False Data Injection (FDI) attacks. Therefore, they proposed a DBN-based method to identify FDI attacks; this method significantly improves the detection accuracy of FDI attacks. Moreover, the work~\cite{8233155} proposed a wide and deep method to detect electricity thieves in power grids. In addition to attack detection, privacy protection is also crucial. Therefore, DeepCoin~\cite{ferrag2019deepcoin} proposed RNN combined with blockchain technology to simultaneously identify attacks and protect data privacy. Moreover, Yin et al.~\cite{yin2017deep} also used a RNN model to identify the types of network attacks and the experimental results verified its good performance. Despite the advances of DL models, they often have poor generalization ability. The authors~\cite{zhang2019intrusion} proposed a deep belief network combined with improved genetic algorithms to improve generalization ability.

In Internet of medical things and healthcare systems, the shared and uploaded medical data are usually encrypted to protect the patient's data privacy in medical data. In~\cite{guo2020privacy}, the authors proposed a data-search method based on encrypted images; this method improves the data-search accuracy while preserving data privacy.

In the training process of Distributed DL (DDL), each node needs to share original data or parameters; this process may be vulnerable to network attacks and privacy leakage. In~\cite{aono2017privacy}, the authors combined the original architecture with an additive homomorphic encryption algorithm to ensure privacy and security of the data. Meanwhile, Liu et al.~\cite{liu2020padl} proposed a layer-wise importance propagation algorithm to improve the privacy protection. Li et al.~\cite{li2020toward} proposed a secure and privacy-preserving DDL that combines encryption and signature algorithms to ensure data security. Moreover, Saharkhizan et al.~\cite{Saharkhizan_An_2020} introduced a cyber-attack detection method for IoT devices and systems. They utilized a decision tree to merge a set of LSTM modules and get an aggregated result at the final stage.

\subsection{Upper Applications}

\begin{figure*}[t] 
    \centering 
    \includegraphics[width=0.78\textwidth]{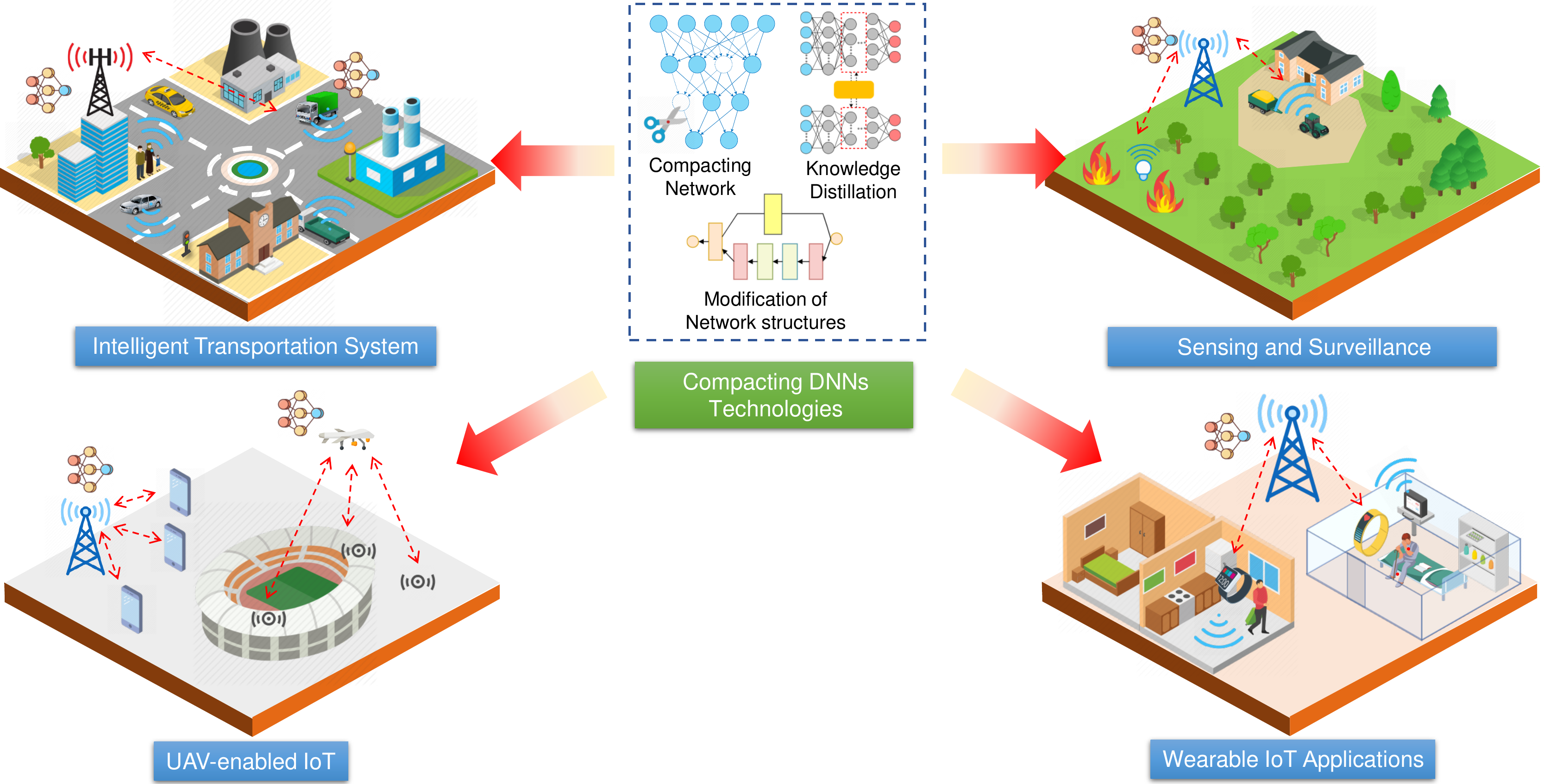} 
    \caption{Upper applications of compacting-DNNs technologies in IoT.} 
    \label{fig:scenario} 
\end{figure*}

The upper applications can be categorized into Intelligent Transportation Systems (ITS), UAV-enabled IoT applications, environmental sensing and surveillance, and wearable IoT applications, as shown in Fig.~\ref{fig:scenario}.

\subsubsection{Intelligent Transportation Systems (ITS)}

The proliferation of diverse IoT nodes in ITS leads to massive ITS data, which can be used to detect traffic bottlenecks, identify traffic jams, optimize traffic control. Due to the resource limitation of IoT nodes, ITS data has often been outsourced to remote clouds, which cannot fulfill the latency and context-aware demands of ITS applications. Thus, the advent of mobile edge computing can offload the computing tasks to nearby edge nodes, which nevertheless cannot process computational-intensive tasks, like DL algorithms. 

There are a few attempts towards the solutions to these issues. In particular, this work~\cite{jhzhou:ACM-TIST2019} presented a composite CNN structure, which combined factorization convolutional layers and compression layers. This new CNN model could significantly reduce the model size and decrease the computation costs with competitive accuracy in the classification task of traffic signs. Realistic experimental results on Jetson TX2 (an edge device developed by NVidia) have further verified the effectiveness of the proposed model. Meanwhile, pruning is highly versatile and has a wide range of applications including real-time target detection and real-time drone applications. Before deploying a model~\cite{noauthor_bringing_nodate}, pruning operations are generally required to reduce the size and the model's inference time. Jetbot is a self-driving car model with Jetson Nano as the core of its calculation. Jetson Nano's volume is very small, but it can provide 470 GFLOPS computing power. Pruner which utilizes the pruning method they mentioned above to lightweight the model is a key procedure in the whole process. Consequently, the pruner reduced the size of the model to achieve real-time detection. Impressively this model can reach 60 Frames Per Second (FPS) in the inference phase.

In ITS, autonomous driving has received extensive attention recently. Driver gesture recognition is a key component of advanced diver assistance systems, which has potential value in applications such as autonomous driving, driver behavior understanding, human-computer interaction, and driver attention analysis. Liu et al.~\cite{liu_driver_2019} designed a novel network RM–ThinNet, which uses a lightweight model to estimate the driver’s posture. Extensive experimental results also demonstrate the effectiveness of the proposed model.  In~\cite{yu_short_2019}, the authors used LSTM to build a short-term traffic flow prediction model based on the driving data of private cars and minibusses. In~\cite{tian2020traffic}, the LSTM is not only used to extract time-series information but also is combined with stack auto-encoder to extract spatial information in traffic data, thereby achieving more accurate traffic flow prediction.

\subsubsection{UAV-enabled IoT Applications}

Unmanned Aerial Vehicle (UAV) hardware has weak computing power hence needs fewer algorithm parameters, less memory, and short inference time to achieve real-time target detection. Traditional solutions are not quite suited for UAV scenarios. Zhang et al.~\cite{zhang_slimyolov3_2019} utilized the channel pruning method to deal with this problem. Concretely they pruned the YOLOv3 model to learn an efficient deep target detector through channel pruning. In order to enhance the channel-level sparsity, they used L1 regularization on the channel scale factor. In this way, they obtain a ``slim'' object detector with a small scale factor. The final results showed that SlimYOLOv3 is faster and better than YOLOv3 in real-time drone applications.

Matching aerial images with actual road landmarks (also known as ``air road matching'') is a key technology to enhance the navigation of Unmanned Aerial Systems (UAS) in GPS-denied urban environments. Considering that UAS usually has limited computing power and storage space, the authors~\cite{zhao_a_2019} propose a lightweight CNN architecture for cross-domain air road matching. In particular, this neural network structure has two branches, each of which uses CNN as the feature extraction module. Each branch uses MobileNetV2 block as the basic block and partially shares the weights between each other.

The quality of images taken by drones is easily affected by weather and ambient light. In order to obtain a better quality of remote sensing images, Wu et al.~\cite{wu_silmrgbd_2020} designed a model namely SlimRGBD, which enables drones to automatically implement denoising operations when they took unclear photos. They trained a GAN model in advance to generate enough noisy images for the drone to learn the noise distribution. They then applied a structure similar to ResNet to achieve the denoising effect for noisy images. However, both these two networks are relatively bulky and complex so that they may not be suitable for UAVs with limited resources. Therefore, the authors used channel pruning to reduce the size of the networks so that the models are more portable for UAV scenarios.

It is challenging to achieve the autonomous navigation and control of UAVs. Baldini et al.~\cite{baldini_learning_2020} proposed a lightweight learning model for autonomous navigation and landing of UAVs. In particular, the proposed framework consists of a new online pose estimation approach which consists of a CNN for image regression and two different sizes of LSTMs for position and orientation estimations. The CNN module named ResNet18, utilized an attention mechanism as the same as SE module and shortcut connections. This new approach boosts a 25\% improvement compared to its counterparts in estimation accuracy.

There are trade-offs between classifier accuracy and computation complexity, considering limited resources UAVs. Rajagopal et al.~\cite{rajagopal_a_2020} presented a novel and effective model based on a multi-objective optimization method for scene recognition. Concretely, the proposed approach allows the UAV to capture frames. Meanwhile, an optimal multi-objective particle swarm optimization is used to a CNN to derive a model. This novel method achieved the highest accuracy and lowest computation time compared to other counterparts. Kyrkou et al.~\cite{kyrkou_E_2020} came up with a lightweight CNN called EmergencyNet for drone-based emergency monitoring. EmergencyNet that utilizes atrous convolutions to capture multi-resolution features is capable of being efficiently executed at UAVs, consequently achieving higher performance compared to existing models with minimal memory requirements.

\subsubsection{Environmental Sensing and Surveillance}

The lightweight DNNs can be used for IoT, which has been deployed for environmental sensing and surveillance. Huynh et al.~\cite{huynh_deepmon_2017} developed a framework called DeepMon to optimize diverse CNNs on mobile GPUs and proposed an efficient and energy-saving DL inference system for mobile devices. First, they leveraged the similarity between consecutive frames of the video to design an intelligent caching mechanism for the convolutional layer. Second, they accelerated the high-dimensional matrix multiplication in the convolutional layer through matrix decomposition, which is the bottleneck in running convolutional layers on GPUs. DeepMon uses the adjusted Tucker-2~\cite{kim_compression_2015} decomposition to further reduce the cut-down time of the convolutional layer.

Regarding sensing applications, Yao et al.~\cite{yao_deepiot_2017} developed a new general compression method for neural networks after absorbing the idea of dropout to drop hidden elements. Specifically, the redundant nodes are deleted as much as possible to reduce the size of the network after finding the optimal dropout probability of each hidden element in the network. Moreover, this framework also compresses sparse neural networks into dense structures with smaller sizes. Furthermore, the proposed model can be directly used in IoT devices without extra modifications. Extensive experiments also verify the effectiveness of the proposed framework.

Moreover, lightweight DNNs can also be used in other sensing and surveillance scenarios. In particular, Wang et al.~\cite{wang_lightweight_2019} combined the structure of the inverted residuals in MobileNetV2 and the SE module in SENet to design a lightweight neural network for weather monitoring. This model can effectively reduce the memory cost. Moreover, Yang et al.~\cite{yang_non-temporal_2019} combined the depth-wise separable convolution unit and SE Module to design a lightweight neural network for fire detection. Since fire detection is mainly based on color and different color channels have different sensitivity levels, the authors introduced a channel multiplier to emphasize the color channels to get a better detection result. 

Zhao et al.~\cite{Zhao_L_2020} proposed a DL technique for intelligent-edge surveillance. They utilized depth-wise separable convolution to reduce computational cost and combine edge computing with cloud computing to reduce network traffic. This method reaches 16 frames per second at the edge device and the cost at the edge device is only one-tenth of that of the centralized server. Experiments show the convincing results of the proposed technique in terms of computational cost and accuracy. To achieve good accuracy and comparative precision, Yassine et al.~\cite{Yassine_H_2020} adopted MobileNet with transfer-learning approach to construct a DL model for human detection for video surveillance.

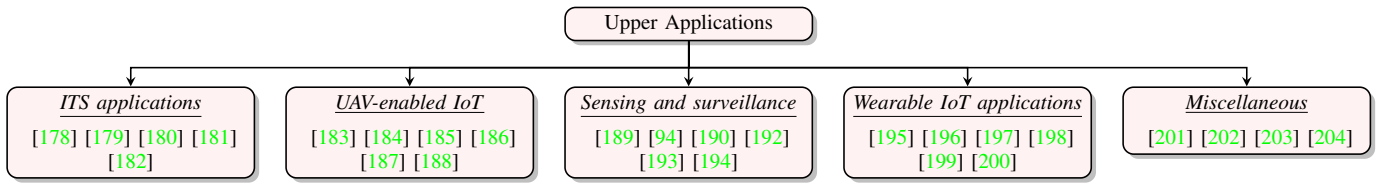
\begin{figure*}[t]
		\centering
		\footnotesize
		\subfigure{
		\begin{forest}
				for tree={              
					draw, semithick, rounded corners,
					text width = 31mm, text badly centered,
					edge = {draw, semithick, -stealth},
					anchor = north,
					grow = south,
					forked edge,            
					s sep = 4mm,    
					l sep = 6mm,    
					fork sep = 3.5mm,  
					tier/.option=level,
					fill=red!5,
				}
				[Upper Applications
				[\underline{\textit{ITS applications}}\\	 \vspace{0.15cm} \cite{jhzhou:ACM-TIST2019}\cite{noauthor_bringing_nodate}\cite{liu_driver_2019}\cite{yu_short_2019}\cite{tian2020traffic}]
				[\underline{\textit{UAV-enabled IoT}} \\ \vspace{0.15cm}
			      \cite{zhang_slimyolov3_2019}\cite{zhao_a_2019}\cite{wu_silmrgbd_2020}\cite{baldini_learning_2020}\cite{rajagopal_a_2020}\cite{kyrkou_E_2020} 
				]
				[\underline{\textit{Sensing and surveillance}} \\ \vspace{0.15cm}
			     \cite{huynh_deepmon_2017}\cite{kim_compression_2015}\cite{yao_deepiot_2017}\cite{yang_non-temporal_2019}\cite{Zhao_L_2020}\cite{Yassine_H_2020} 
				]
				[\underline{\textit{Wearable IoT applications}} \\ \vspace{0.15cm}
			     \cite{sourav_association_2016}\cite{richoz_transportation_2020}\cite{loh_low_2020}\cite{lu_one-shot_2019}\cite{Nazari_M_2020}\cite{Tang_L_2021}
				]
				[\underline{\textit{Miscellaneous}} \\ \vspace{0.15cm} \cite{bhardwaj_memory_2019}\cite{han_ese_2017}\cite{thakker_compressing_2019}\cite{jahromi_an_2020}
				]
				]	
        \end{forest}
        }
		\caption{Summary of compacting-DNNs technologies in IoT applications}
		\label{fig6-1}
\vspace*{-0.5cm}
\end{figure*}

\subsubsection{Wearable IoT Applications}
DL methods can also be used for wearable IoT devices. However, it is also challenging to deploy DNNs at the limited resources of IoT devices. There are several studies working on designing portable DL models for wearable IoT devices. In particular, Bhattacharya et al.~\cite{sourav_association_2016} devised a framework namely SparseSep, which leverages the sparsity of the fully connected layer and the separation of the convolution kernel, reaching the goal of no retraining, no cloud offloading, a low-resource platform, and minimizing model changes. The framework SparseSep consists of a layer compression compiler, a runtime framework called sparse inference runtime, and a separator called convolutional separation runtime. Extensive experiments were conducted on diverse embedded platforms such as Qualcomm Snapdragon, ARM Cortex, and Nvidia Tegra to further demonstrate the effectiveness of SparseSep.

Meanwhile, Richoz et al.~\cite{richoz_transportation_2020} introduced three independent lightweight DNNs into wearable devices, corresponding to three types of sensors, namely motion sensors, sound sensors, and visual sensors. On this basis, the multimodal fusion methods are used to better recognize the transportation pattern. Moreover, with the help of wearable devices, Loh et al.~\cite{loh_low_2020} classified the collected electrocardiogram signals for cardiac arrhythmia detection. Specifically, discrete wavelet transform and CNN were used to classify the collected signals. Through the quantization and pruning of CNN, a certain amount of parameters are reduced. Furthermore, Lu et al.~\cite{lu_one-shot_2019} proposed a lightweight I3D-based network for gesture recognition. This model has spatio-temporal separable 3D convolution and fire modules, which can effectively extract discriminative features.

ElectroEncephaloGraphy (EEG) classification is essential to be deployed in wearable IoT devices. Complex LSTMs that have been widely used in sequential applications can hardly be deployed at wearable devices due to computations and memory requirements. Nazari et al.~\cite{Nazari_M_2020} proposed a multi-level binarized LSTM by introducing binary LSTMs to cope with these problems. By using quantization methods, this algorithm brings performance efficiency with EEG classification to wearable IoT applications. For human activity recognition using wearable sensors, Tang et al.~\cite{Tang_L_2021} proposed a lightweight CNN with a set of low dimensional Lego filters stacking for convolutional filters. The proposed CNN can greatly reduce computation cost and memory cost while it is faster and more accurate than its counterparts. 

\subsubsection{Miscellaneous}

Portable DNNs have been used in other distributed IoT scenarios. Model compression is an important research area for deploying DL models on the IoT. However, due to extreme memory limitations, even the compressed model cannot be accommodated by a single device. Therefore, researchers considered distributing the model on multiple devices, but the main problem is the cost of memory and communication overhead. 

In particular, Bhardwaj et al.~\cite{bhardwaj_memory_2019} proposed network of neural networks, which compresses a large, pre-trained ``teacher'' network into several separate and compact ``student'' modules without accuracy loss. They also proposed a knowledge distribution algorithm based on network science, allowing the teacher model to traverse each student network on the generated disjoint partitions. Moreover, Han et al.~\cite{han_ese_2017} proposed the efficient speech recognition engine, which can be used to speed up the prediction of LSTM. The pruning method used is a perceptible method of load balancing, which makes the pruned model easy to process in parallel. And in the weight quantization stage, analyze the weight dynamic range of all matrices in each LSTM layer to avoid data overflow. Thakker et al.~\cite{thakker_compressing_2019} proposed a method to compress RNN by using Kronecker Product (KP). The authors also proposed hybrid KP RNNs to reduce the accuracy loss caused by using KP. As a result, hybrid KP divides the matrix into an unconstrained upper part and a lower part. Jahromi et al.~\cite{jahromi_an_2020} proposed an enhanced stacked LSTM method for malware detection in IoT.

\subsection*{\bf Summary and Insights}

DL methods have been widely used in IoT because of their powerful capabilities of data processing. Fig.~\ref{fig6-1} summarizes the above upper applications of compacting-DNNs technologies. We find that DNN deployment methods vary with the diversity of IoT devices. Meanwhile, the compression of DNN is also significant, especially at ultra-small IoT devices. 
When compacting-DNNs technologies are applied in IoT scenarios, they should be adapted to fulfill the characteristics of IoT devices. Moreover, there is a trade-off between performance and resources.

\section{Future Directions}
\label{sec:future}


The future directions of compacting DNNs can be categorized into four major directions: exploration of model compression techniques, automated model compression methods, deployments of compacted DNNs in IoT, and Blockchain as well as deep learning for IoT.



\subsection{Exploration of model compression techniques}

New model compression techniques are expected to further reduce the model size despite several recent attempts. For example, Guo et al.~\cite{guo_single_2019} proposed a single-path one-shot model, which has a simplified structure, in which feature maps can be generated with cheap operations. Meanwhile, the authors~\cite{lin_towards_2019} pointed out that traditional network pruning discards invalid filters, so they utilized filter grafting to reactivate them from the perspective of improving accuracy. On the other hand, other studies attempt to introduce new activation functions so as to improve the convergence and enhance the computing efficiency. For example, Zoph et al.~\cite{zoph_learning_2018} proposed a new activation function $f(x)=x \tanh \left(e^{x}\right)$, which can achieve fast convergence and can help speed up the model training process. Starting from the model optimization method, Ma et al.~\cite{ma_a_2019} proposed a Bayesian optimization framework to replace the traditional method of error guiding model optimization and achieved dramatically improved efficiency. Other compression techniques should be further explored in the future.

\subsection{Automated model compression methods}

Along with the trend of Automated ML, model compression is also expected to be automated. In particular, using NAS for acceleration is becoming a new model compression method~\cite{han_ghostnet_2019}. The work~\cite{liu_progressive_2018} adopted NAS as a model search framework that applied supervised learning to help the network find the optimal parameters and produce impressive results. Following this idea, studies~\cite{tan_efficientnet_2019} and~\cite{tan_efficientnet_2019} proposed similar frameworks. On the other hand, GAN has also become another direction in automated model compression. For example, Meng et al.~\cite{meng_filter_2020} used the idea of GAN to let the generator generate the cropped network, and discriminator to determine whether it belongs to the original network or the cropped network so as to perform more effective network structured cropping. Future studies may harness other techniques to achieve automated model compression.

\subsection{Deployment of compacted DNNs in IoT}

In the deployment of compacted DNNs in IoT, there are a number of issues to be solved. On the one hand, partitioning strategies~\cite{jeong_lightweight_2019,lin_cost_2020} were used though, the joint consideration of compacted DNNs with the partitioning strategies is still a problem to be solved. In particular, it is worthwhile to investigate the load-balancing problem caused by the distribution of compacted DNNs deployed at either IoT devices or edge nodes and the integration of input and output generated by each DNN at each computing facility. On the other hand, the communication cost caused by transmitting DNNs models between IoT devices and edge nodes (or clouds) needs to be considered in deployment of compacted DNNs. Although there are some preliminary results such as~\cite{bhardwaj_memory_2019}, which splits the teacher's knowledge into multiple disjoint partitions and consequently let the student modules be disjoint and compressed, there are still a number of issues to be solved in the future.

\subsection{Blockchain and deep learning for IoT}

As a public distributed ledger, blockchain is an ideal complement to IoT with several key characteristics including decentralization, traceability, privacy and security protection. Ferrag et al.~\cite{Ferrag_B_2019} reviewed different blockchain applications for IoT and left some challenging discussions. Moreover, Dai et al.~\cite{Dai_B_2019} presented a novel paradigm called blockchain of things bringing four major merits: 1) interoperability across IoT devices, 2) traceability, 3) reliability of IoT data, 4) autonomic interactions of IoT systems. Despite the advances, many areas remain challenging, such as intrusion detection. From the collected literature, we believe that DL is a powerful tool for data-driven tasks. It is quite promising to integrate blockchain technology with deep learning in the future.

\section{Conclusion}
\label{sec:conc}

This paper presents a state-of-the-art survey on compacting DNNs for IoT scenarios. We first give an overview of DNNs as well as IoT, discuss the challenges in using DNNs in IoT, and elaborate on the fundamentals of compacting-DNNs technologies. We next categorize compacting-DNNs technologies into three types: compacting network model, KD, and modification of network structures. In each type of compacting DNNs approaches, we also elaborate on the exact techniques and implemented models. Moreover, we discuss the applications of compacted DNNs in IoT and point out several future directions in this area. We believe that this comprehensive survey will further foster the wide adoption of DL for IoT.

\appendices
\section{\footnotesize List of Abbreviations}
\label{sec:abbr}

\vspace{0.2cm}

\begin{center}
\footnotesize

\begin{supertabular}{ll}
\textit{Abbreviation} & \textit{Description}   \\

5G           & Fifth Generation                                       \\
AI           & Artificial Intelligence                                \\
ANN          & Artificial Neural Network                              \\
ATN          & Asymmetric Ternary Network                             \\
BCNN         & Binarized CNN                                          \\
CNN          & Convolutional Neural Network                           \\
CP           & Canonical Polyadic                                     \\
CV           & Computer Vision                                        \\
DBN          & Deep Belief Network                                    \\
DDL          & Distributed DL                                         \\
DL           & Deep Learning                                          \\
DNN          & Deep Neural Network                                    \\
EEG          & ElectroEncephaloGraphy                                 \\
EI           & Edge Intelligence                                      \\
FDI          & False Data Injection                                   \\
FLOPS        & FLoating-point Operations Per Second                   \\
FPS          & Frames Per Second                                      \\
GAN          & Generative Adversarial Network                         \\
GPS          & Global Positioning System                              \\
GPU          & Graphics Processing Unit                               \\
GRU          & Gated Recurrent Unit                                   \\
IaaS         & Infrastructure-as-a-Service                            \\
IoT          & Internet of Things                                     \\
IoV          & Internet of Vehicles                                   \\
ITS          & Intelligent Transportation Systems                     \\
KD           & Knowledge Distillation                                 \\
KP           & Kronecker Product                                      \\
LSTM         & Long Short-Term Memory                                 \\
ML           & Machine Learning                                       \\
NAS          & Network Architecture Search                            \\
PaaS         & Platform-as-a-Service                                  \\
PQ           & Product Quantization                                   \\
QNN          & Quantized Neural Network                               \\
RFID         & Radio Frequency Identification                         \\
ReLU         & Rectified Linear Units                                 \\
RNN          & Recurrent Neural Network                               \\
RQ           & Residual Quantization                                  \\
SaaS         & Software-as-a-Service                                  \\
SE           & Squeeze-and-Excitation                                 \\
SVD          & Singular Value Decomposition                           \\
TPU          & Tensor Processing Unit                                 \\
TWN          & Three elements Weight Network                          \\
UAS          & Unmanned Aerial Systems                                \\
UAV          & Unmanned Aerial Vehicle                                \\
VGG          & Visual Geometry Group                                  \\  
\end{supertabular}
\end{center}

\bibliography{revbib.bib}
\vspace{-15 mm}
\begin{IEEEbiography}[{\includegraphics[width=1in,height=1.25in,clip,keepaspectratio]{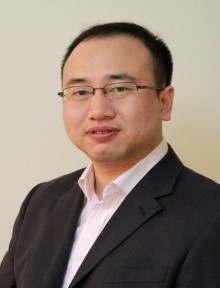}}]{Ke Zhang} (M'09) received the B.S. and M.S. degrees from the University of Electronic Science and Technology of China (UESTC) in 2002 and 2006, respectively, and the Ph.D. degree from the School of Computer Science and Engineering, UESTC, in 2010. He is currently an Associate Professor with the School of Computer Science and Engineering, UESTC, Chengdu, China. Ke Zhang has published more than 60 research articles in many journals and conferences. He is also the reviewer of many prestigious international conferences and journals, and he owns more than 14 national invention patents. His research interests include the internet of things, AI, big data, computer networks, and data fusion. His scientific research at UESTC started with designing an intelligent firewall system based on an Intel IXA network processor. Much of his focus has been on sensor networks, AI, and big data since 2006, participated in several research projects, such as the National Natural Science Foundation of China, the Sichuan Science and Technology Program, and Aviation Science Fund Projects and so on. He is also a Senior Member of the China Computer Federation (CCF) and the CCF's Member of the committee of experts on computer applications.
\end{IEEEbiography}
\vspace{-15 mm}
\begin{IEEEbiography}[{\includegraphics[width=1in,height=1.25in,clip,keepaspectratio]{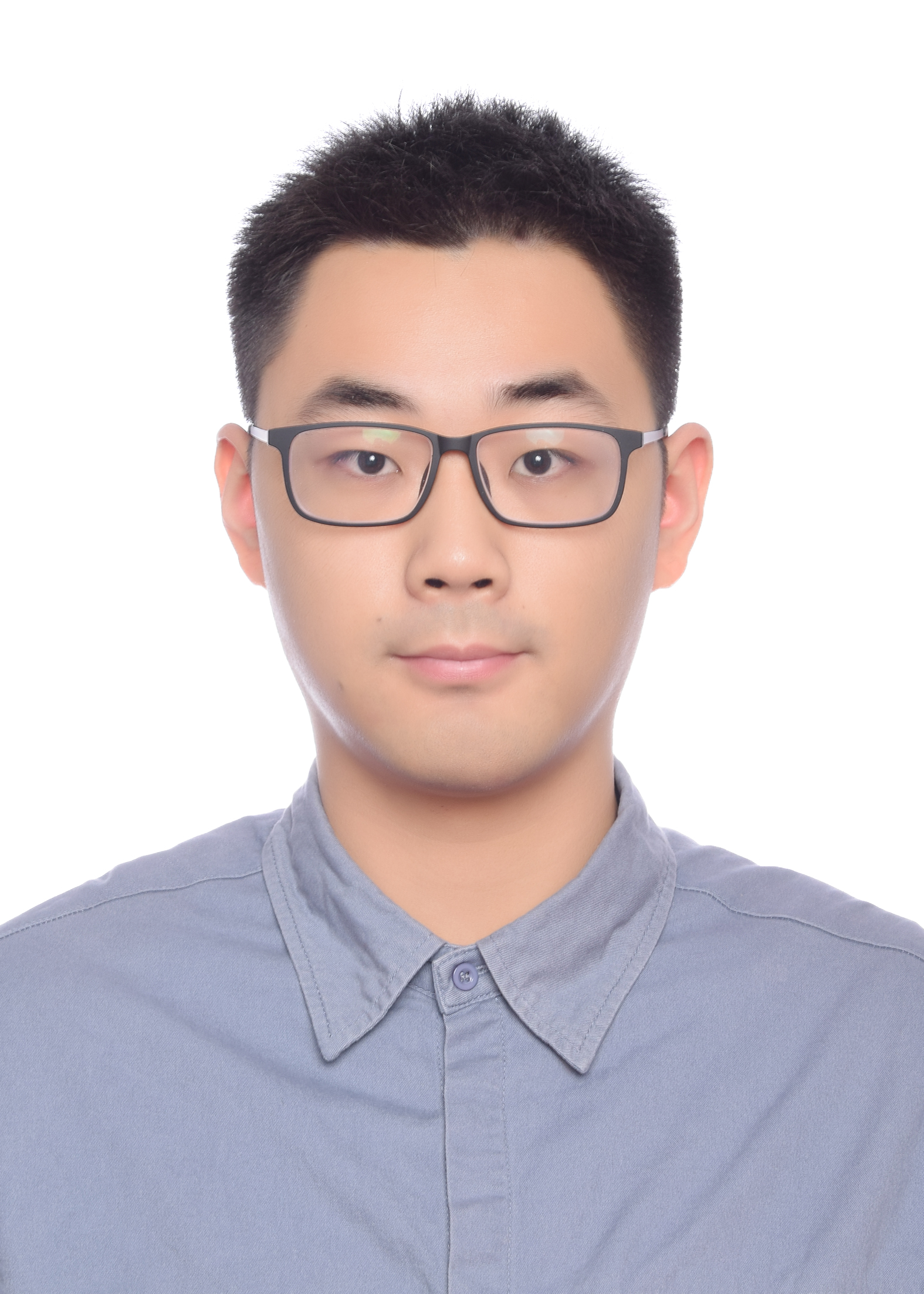}}]{Hanbo Ying} received the B.E. degree in network engineering from the University of Electronic Science and Technology of China (UESTC), Chengdu, China, in 2018. And now he is currently pursuing the M.A.Eng degree in Information and Communication at UESTC. His current research interests include big data, compression and acceleration in neural networks and internet of things.
\end{IEEEbiography}
\vspace{-15 mm}
\begin{IEEEbiography}[{\includegraphics[width=1in,height=1.25in,clip,keepaspectratio]{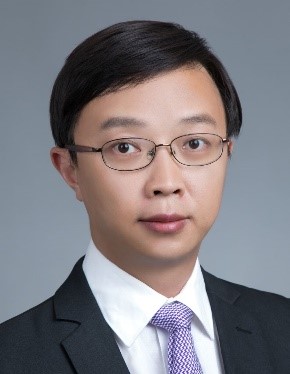}}]{Hong-Ning Dai} (SM'16) is currently with Faculty of Information Technology at Macau University of Science and Technology as an associate professor. He obtained the Ph.D. degree in Computer Science and Engineering from Department of Computer Science and Engineering at the Chinese University of Hong Kong. His current research interests include internet of things big data and blockchain technology. He has served as editors of IEEE Transactions on Industrial Informatics, IEEE Systems Journal, Computer Communications (Elsevier), Connection Science (Taylor \& Francis), and IEEE Access.
\end{IEEEbiography}
\vspace{-15 mm}
\begin{IEEEbiography}[{\includegraphics[width=1in,height=1.25in,clip,keepaspectratio]{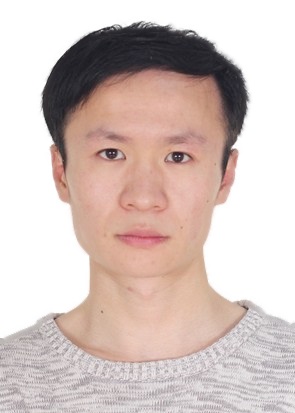}}]{Lin Li} received the B.E. degree in communication engineering from Chongqing University, in 2017. He is currently pursuing the M.A.Eng. degree with the University of Electronic Science and Technology of China (UESTC). His research interests include computer vision, GAN and the internet of things.
\end{IEEEbiography}
\vspace{-15 mm}
\begin{IEEEbiography}[{\includegraphics[width=1in,height=1.25in,clip,keepaspectratio]{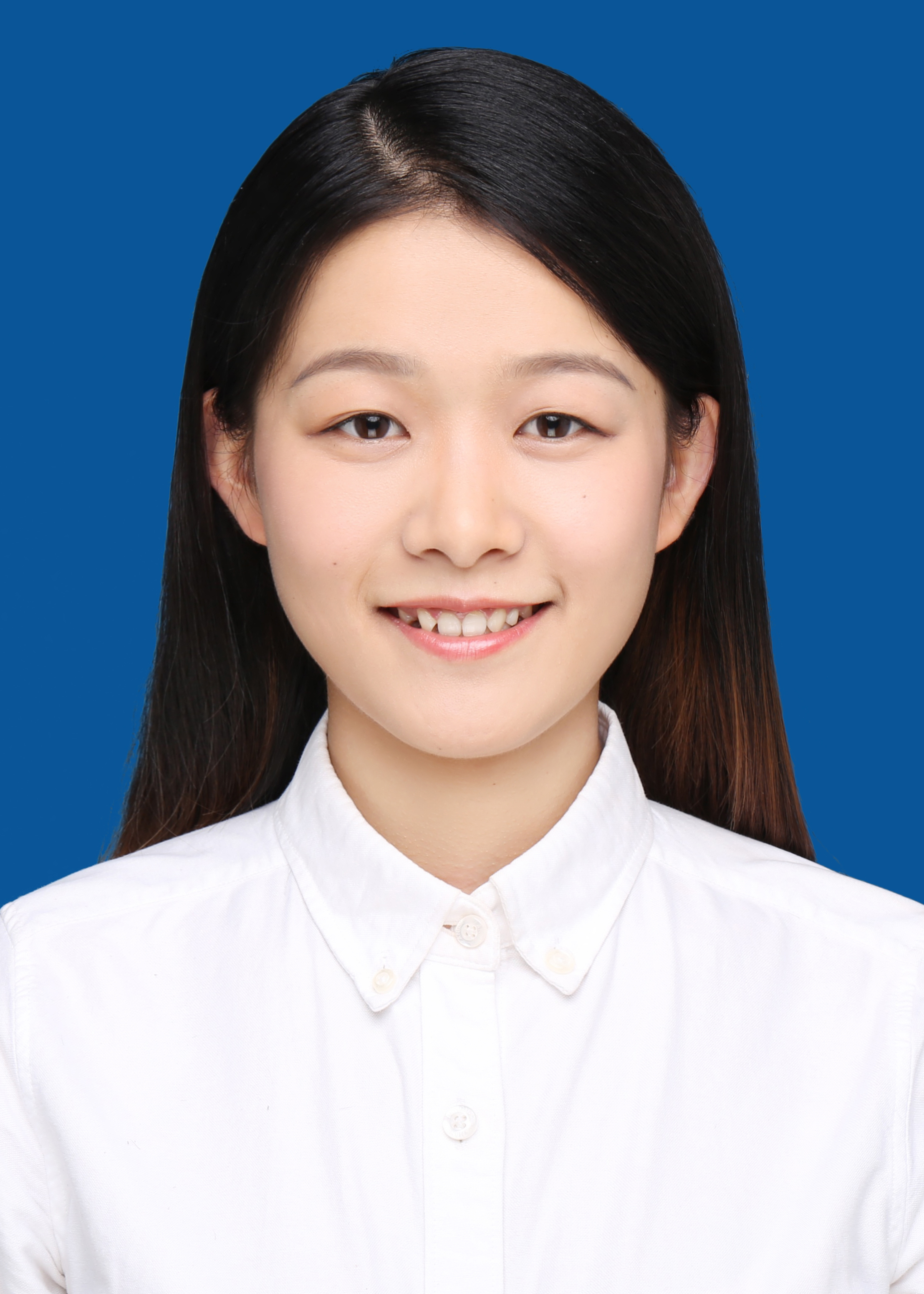}}]{Yuanyuan Peng} received the B.E. degree in Internet of Things engineering from Southwest Petroleum University, in 2019. And now she is pursuing the MA.Eng degree in Computer Science and Engeneering at University of Electronic Science and Technology of China (UESTC). Her research interests include blockchain, IoT, compression and acceleration in neural networks.
\end{IEEEbiography}
\vspace{-15 mm}
\begin{IEEEbiography}[{\includegraphics[width=1in,height=1.25in,clip,keepaspectratio]{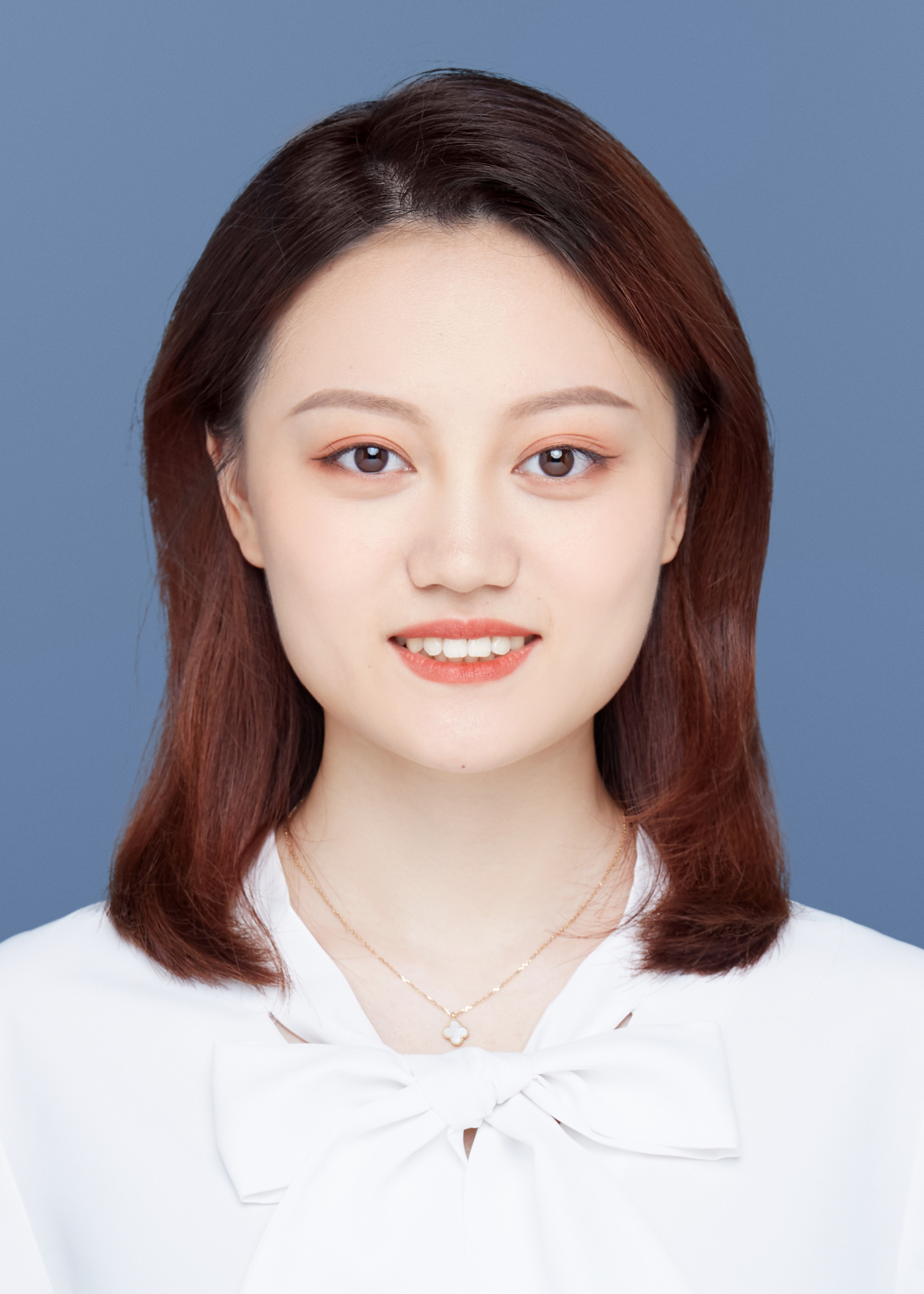}}]{Keyi Guo} is pursuing a B.A. degree in Mathematics at New York University and is expected to graduate in 2021. Her research interests include big data, numerical optimization and the internet of things.
\end{IEEEbiography}
\vspace{-15 mm}
\begin{IEEEbiography}[{\includegraphics[width=1in,height=1.25in,clip,keepaspectratio]{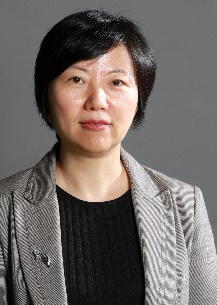}}]{Hongfang Yu} professor at School of Communication and Information Engineering of University of Electronic Science and Technology of China (UESTC), She received her M.S. degree and Ph.D. degree in Communication and Information Engineering in 1999 and 2006 from UESTC, respectively. Her current research interests include data center networking, network (function) virtualization, cloud/edge computing, distributed AI and Blockchain.
\end{IEEEbiography}

\end{document}